\documentclass{article}

\usepackage{PRIMEarxiv}

\usepackage[utf8]{inputenc}
\usepackage[T1]{fontenc}
\usepackage{microtype}
\usepackage{authblk}
\usepackage{graphicx}
\usepackage{amsmath, amssymb}
\usepackage{booktabs}
\usepackage{array}
\usepackage{float}
\usepackage[colorlinks=true,allcolors=blue]{hyperref}
\usepackage[style=numeric,maxnames=3,minnames=1]{biblatex}
\graphicspath{{figures/}}

\title{Knowing When to Stop:\\ Bayesian Optimal Stopping for LLM Evaluations}
\author{\large\textbf{Toby D. Pilditch}}
\affil{\normalsize UK AI Security Institute, London, UK}
\affil{\normalsize \texttt{toby.pilditch@dsit.gov.uk}}

\begin{document}

\maketitle

\begin{abstract}
LLM evaluations often use fixed sampling budgets, testing every item the same number of times even after estimates are precise. We introduce \texttt{optstop}, a precision-based adaptive stopping framework that treats evaluation as a sequential measurement problem: keep sampling where uncertainty remains high, and stop where estimates are precise or stable enough. The framework builds on hierarchical Bayesian inference, supports binary, ordinal, and continuous outcomes, and keeps every benchmark item eligible for sampling, without requiring a calibrated item bank. It runs live or retrospectively, and includes a safeguard that samples more cautiously as measured performance approaches zero, where rare successes matter most. In an illustrative 200-item, 10-epoch evaluation, it removes 57\%--97\% of planned trials across nine validation settings, with overall conclusions equivalent to the full run. These results show that LLM evaluation compute can be allocated by uncertainty rather than by fixed repetition counts, with the magnitude of savings depending on evaluation design.
\end{abstract}

\section{Introduction}

\subsection{Problem Statement}
The evaluation of large language models has become a central pillar of AI safety assurance, with major laboratories, regulatory bodies, and independent auditors conducting increasingly comprehensive testing campaigns to characterise model capabilities and risks prior to deployment (see, e.g., \cite{shevlane2023model,anthropicSystemCards}). Frontier models are now routinely assessed across hundreds of distinct tasks spanning reasoning, factual knowledge, coding, multilingual competence, and safety-relevant behaviours \cite{liang2022holistic,srivastava2023beyond}, with individual benchmarks such as BIG-bench comprising over 200 tasks \cite{srivastava2023beyond} and HELM evaluating models across 42 scenarios \cite{liang2022holistic}.

Each task typically entails hundreds or thousands of test items, and the stochastic nature of language model generation necessitates multiple repetitions per item to obtain stable performance estimates. For a suite of 200 tasks with 1{,}000 items and 5 repetitions each, a single model requires one million inference calls - before accounting for comparisons across model families or evaluation conditions. This computational burden is compounded by the rapid pace of model development: new releases demand fresh evaluation, while meaningful benchmarking often requires re-testing predecessors under identical conditions \cite{kiela2021dynabench,dehghani2021benchmark}. As suites expand to address agentic behaviours \cite{kinniment2023evaluating} and multi-step tasks requiring tool use \cite{mialon2023gaia}, the resource demands grow correspondingly.

Yet despite these costs, current evaluation practices predominantly follow what has been termed the ``highest-number-is-best approach'' \cite{miller2024adding,luettgau2025hibayes}: reporting point estimates of aggregate performance without systematic uncertainty quantification, appropriate treatment of hierarchical data structure, or formal statistical validation \cite{burnell2023rethink}.

This trajectory raises a fundamental methodological question: are current evaluation practices systematically miscalibrated, expending computational resources beyond what statistical inference requires? Or conversely, terminating data collection before reliable conclusions can be drawn? If so, the inefficiency - or the inferential risk - scales with every new model and every expanded benchmark suite.

\subsection{Statistical Sufficiency in Evaluations}
\begin{figure}[htbp]
  \centering
  \includegraphics[width=0.8\linewidth]{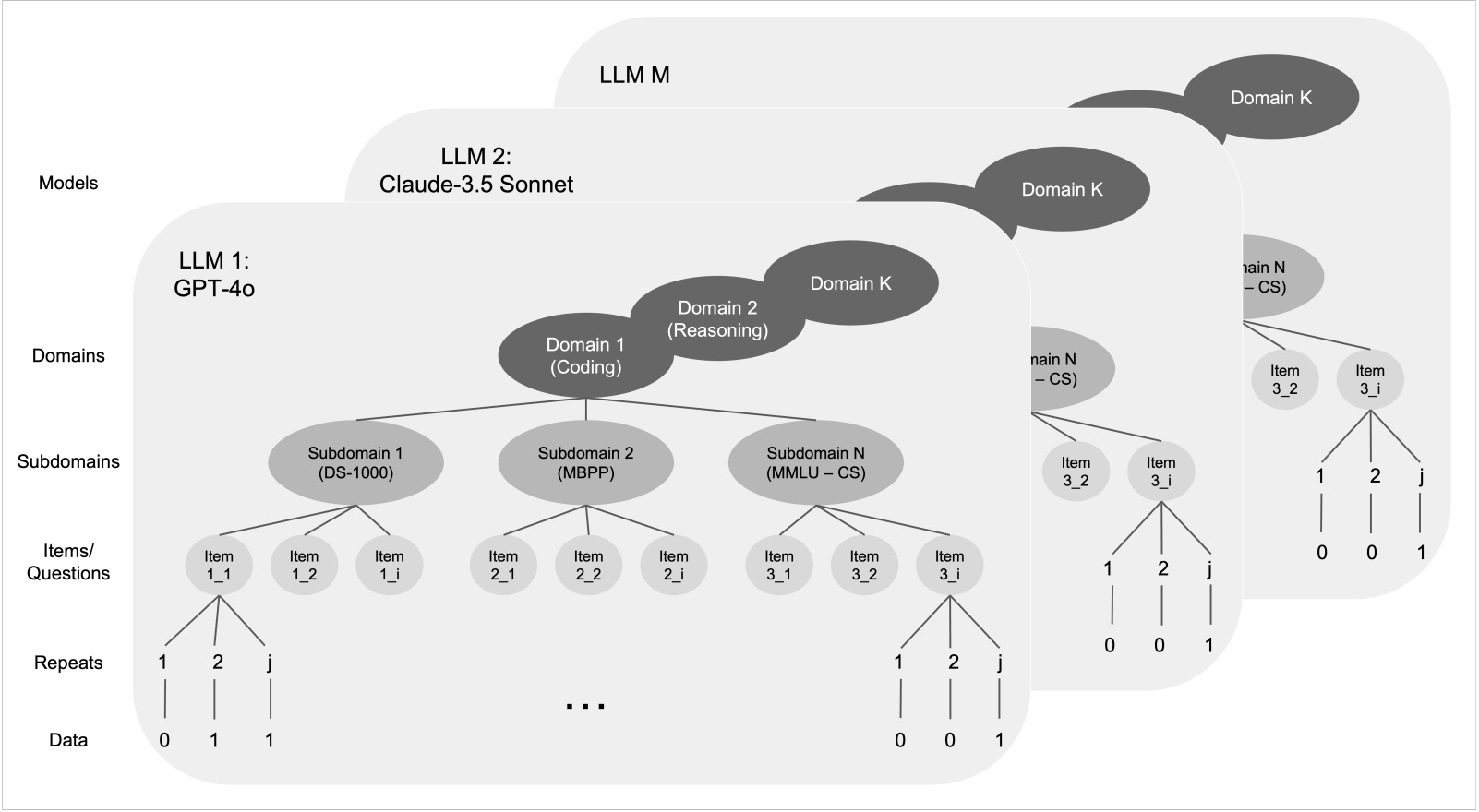}
  \caption{Example hierarchically nested structure of evaluation data. Taken from \cite{luettgau2025hibayes}.}
  \label{fig:llm-struct}
\end{figure}

Understanding why current evaluation practices may be miscalibrated requires examining the statistical structure of LLM assessment and the inferential demands it places on practitioners. Evaluation data possess a naturally nested hierarchy: individual responses are generated for specific items (prompts or problems), which are grouped within tasks or subdomains, which in turn cluster within broader capability domains, with the entire structure replicated across multiple models under assessment \cite{luettgau2025hibayes}, see Figure~\ref{fig:llm-struct}. At the finest grain, repeated queries of the same item can yield different responses, necessitating multiple epochs per item to distinguish capability from sampling noise. This hierarchical structure is pervasive across evaluation contexts yet rarely accorded appropriate statistical treatment.

Compounding this structural complexity is the heterogeneity of performance across the evaluation space. Model capabilities do not degrade or improve uniformly: the ``jagged frontier'' phenomenon - borrowing a metaphor from Dell'Acqua et al.'s study of AI-augmented professional tasks \cite{dell2023navigating} - describes how a model may excel at one capability while failing at a superficially similar one.

Performance varies across tasks, difficulty levels, prompt formulations, and their interactions, and the uncertainty in any evaluation is itself uneven: some model-task combinations yield consistent results after few observations, while others exhibit high variance that demands extended sampling to resolve. Moreover, current practice applies uniform sampling regimes regardless of inferential goal or intrinsic variability.

Conventional frequentist approaches - sample means with standard errors or confidence intervals - presuppose large samples, approximate normality, and variance homogeneity, assumptions routinely violated in evaluation contexts with sparse, non-continuous, or heterogeneous data \cite{luettgau2025hibayes,miller2024adding}. Hierarchical Bayesian models address these limitations by explicitly representing the nested structure of evaluation data, enabling partial pooling across levels \cite{gelman2007data}. Posterior credible intervals provide valid uncertainty quantification even in low-data regimes, without asymptotic assumptions. The HiBayES framework \cite{luettgau2025hibayes} has demonstrated these advantages for LLM evaluation specifically, showing more calibrated uncertainty estimates and more robust inferences than conventional methods.

This paper introduces \texttt{optstop}, a precision-based optimal stopping framework for LLM evaluation that similarly leverages hierarchical Bayesian principles. Where frameworks such as HiBayES address the analysis of evaluation data after collection - answering \emph{what does this data tell us?} - \texttt{optstop} addresses the logically prior question: \emph{when can data collection safely stop?} The package implements adaptive sequential stopping rules that monitor the width of Bayesian credible intervals during evaluation, automatically terminating data collection for individual items and model-task groupings once estimates reach a user-specified precision.

\subsection{The Value of Efficiency in Testing}

The stopping question reduces to a resource allocation problem: how to maximise the expected information gain from the next observation. As posterior uncertainty decreases non-uniformly across the evaluation space, an efficient strategy terminates collection where precision is adequate and concentrates effort where uncertainty persists.

This principle has immediate practical consequences. The direct costs of comprehensive evaluation are substantial - single tasks in some agentic benchmarks consume tokens worth hundreds of US dollars \cite{chollet2024o3,luettgau2025hibayes} - and these costs shape what gets evaluated, forcing trade-offs between breadth of coverage and depth of assessment. Evaluation latency constrains safety assurance cycles, compressing the window for remediation and deliberation \cite{shevlane2023model}. Beyond cost and time savings, efficiency has implications for environmental sustainability \cite{patterson2021carbon} and equitable access to rigorous evaluation. Most consequentially, adaptive stopping transforms evaluation from a fixed-budget exercise into one of targeted reallocation: computational effort freed from well-characterised model-task combinations can be redirected toward capability boundaries and rare-event scenarios where uncertainty - and safety relevance - is greatest.

\subsection{Related Work}

Several complementary strategies for improving LLM evaluation efficiency have emerged. \emph{Adaptive item selection} methods, grounded in Item Response Theory (IRT) and Computerized Adaptive Testing (CAT), reduce evaluation cost by selecting maximally informative test items for each model. Perlitz et al. \cite{perlitz2024efficient} demonstrate that reliable benchmark rankings can be obtained from a fraction of evaluation items. Hofmann et al. \cite{hofmann2025fluid} combine IRT-based ability estimation with dynamic item selection, and Li et al. \cite{li2025atlas} use Fisher information-guided item selection to reduce required items by up to 90\%. Balk{\i}r et al. \cite{balkir2026confident} extend adaptive testing to continuous scores with precision-based stopping. These approaches reduce the \emph{breadth} of evaluation (fewer items per model) but typically require a pre-calibrated item bank, which may not exist for novel benchmarks or safety-critical tasks where comprehensive coverage is required.

A separate line of work applies optimal stopping to \emph{inference-time} sampling: determining when to stop generating candidate responses for a given prompt. Wan et al. \cite{wan2025beacon} use Bayesian sequential search, while Kalayci et al. \cite{kalayci2025optimal} apply Pandora's Box models. These optimise generation quality, not evaluation design.

The present work occupies a distinct niche: automated, real-time stopping decisions \emph{during} evaluation runs, determining when collected data are sufficient for reliable inference at both the item and grouping levels, without requiring pre-calibrated item parameters or modifications to evaluation content.

\section{The \texttt{optstop} Framework}
\label{sec:framework}

The \texttt{optstop} package is a Python library, publicly available as a GitHub repository (\url{https://github.com/UKGovernmentBEIS/optstop}). It integrates with the \texttt{inspect\_ai} evaluation framework~\cite{ukgovernmentbeis2024inspect} as a live early stopping protocol, and also functions as a standalone tool for retrospective analysis of completed datasets.

\subsection{Precision-Based Stopping}

The core stopping criterion evaluates whether the width of the posterior credible interval for a performance parameter has fallen below a user-specified precision threshold $\delta$. Let $W = \theta_U - \theta_L$ denote the width of the $(1-\alpha)$ credible interval for parameter $\theta$; stopping occurs when $W < \delta$. The framework operates at two levels: one threshold governs stopping at the individual item level (when sufficient repeated evaluations have been collected for a given prompt), and a second governs stopping at the grouping level (when sufficient items have been evaluated for a given model-task combination). A threshold of $\delta = 0.05$ implies precision to within $\pm 2.5$ percentage points at the specified credibility level (default 97\%).

A secondary \textit{stabilisation criterion} detects when further data collection yields diminishing inferential returns. This criterion monitors the slope of recent CI width values within a sliding window, declaring stabilisation when the slope is near zero and not trending toward steeper descent. It serves as a fallback for settings where the width threshold would require impractically many observations (see Appendix~\ref{app:stabilisation} for formal specification).

Valid application requires two conditions: exchangeability of observations within each grouping (their joint distribution should be invariant to permutation), and randomised item presentation order. The \texttt{inspect\_ai} framework supports the latter through its \texttt{sample\_shuffle} option (not enabled by default, but recommended as a precautionary measure when the evaluator cannot guarantee that item ordering is independent of difficulty or other confounds). Evaluators using \texttt{optstop} independently should ensure equivalent randomisation.

\subsection{Conservatism}
\label{sec:conservatism}

A systematic risk attends any early stopping procedure: premature termination before rare but important events have been observed. In LLM evaluation this risk is most acute for capability detection - a model that succeeds on only 1\% of attempts warrants substantially more cautious stopping than one succeeding 99\% of the time. This concern connects directly to the pass@k evaluation paradigm~\cite{chen2021evaluating}, where even a single success across $k$ attempts demonstrates qualitatively different capability than consistent failure.

% Float declared here (within Section 2.2) so that, as a bottom float, it lands on
% the same page as Section 2.4, which introduces and references it.
\begin{table}[!b]
\centering
\caption{Summary of the three inference pathways. $\mu_0 = 0$ by default; the prior scales shown are the pathway defaults (both configurable via \texttt{prior\_mu} and \texttt{prior\_sigma}); $\mu_{\text{group}}$, $\sigma_{\text{group}}$, and $\phi_{\text{group}}$ denote the group-level mean, between-item scale, and precision parameters respectively. Stopping thresholds apply at both item and grouping levels, and the asymmetric conservatism adjustment (Section~\ref{sec:conservatism}) applies to all three pathways when estimated performance falls below the low-performance threshold. Full specifications - including item-level priors, cutpoint parameterisation, and the priors of the Dirichlet-Multinomial fallback model - are given in Appendix~\ref{app:pathways}.}
\label{tab:pathway_summary}
\small
\begin{tabular}{@{}>{\raggedright\arraybackslash}p{0.95in}>{\raggedright\arraybackslash}p{1.55in}>{\raggedright\arraybackslash}p{1.75in}>{\raggedright\arraybackslash}p{1.6in}@{}}
\toprule
 & \textbf{Binary} & \textbf{Ordinal} & \textbf{Continuous bounded} \\
\midrule
Scores & 0/1 (correct/incorrect) & Ordered categories $0, \ldots, K{-}1$ (e.g., 0--10 rubric, $K = 11$) & Bounded continuous, normalised to $[0, 1]$ \\
\addlinespace
Routing & Default pathway & \texttt{ordinal\_tasks} name matching (\texttt{inspect\_ai} and standalone) & \texttt{score\_agg} aggregation (\texttt{inspect\_ai}) or \texttt{continuous\_tasks} name matching (standalone) \\
\addlinespace
Item-level model & Adaptive Beta (data-dependent prior) & Bayesian bootstrap over the modal category & Beta via method-of-moments \\
\addlinespace
Grouping-level model & Logit-normal hierarchical & Hierarchical ordered logistic (cumulative link); Dirichlet-Multinomial fallback & Logit-normal hierarchical on item-level means \\
\addlinespace
Group-level priors & $\mu_{\text{group}} \sim \text{Normal}(\mu_0, 1.5)$; $\sigma_{\text{group}} \sim \text{Exponential}(1)$ & $\mu_{\text{group}} \sim \text{Normal}(\mu_0, 2)$; $\sigma_{\text{group}} \sim \text{Exponential}(1)$ & $\mu_{\text{group}} \sim \text{Normal}(\mu_0, 1.5)$; $\sigma_{\text{group}} \sim \text{Exponential}(1)$; $\phi_{\text{group}} \sim \text{Gamma}(2, 1)$ \\
\addlinespace
Estimand & Population mean success probability & Modal category of the item-averaged distribution & Population mean of item-level means \\
\addlinespace
Stopping criteria & CI width $< \delta$; stabilisation criterion as fallback & Hybrid: modal CI width $< \delta$ gated on low entropy (Pathway~1), or entropy CI width below convergence threshold (Pathway~2) & CI width $< \delta$ on the normalised scale; stabilisation criterion as fallback \\
\bottomrule
\end{tabular}
\end{table}

The package addresses this through an asymmetric conservatism adjustment applied when estimated performance falls below a low threshold (default 1\%): the effective credible interval width used for the stopping comparison is multiplied by a conservatism factor (default $c = 5$), and the slope threshold for the stabilisation criterion is tightened by the same factor. This delays stopping for low-performing combinations, providing additional opportunity for rare successes to be observed, without penalising evaluations where performance is clearly high. A sensitivity analysis of this mechanism across calibrated performance levels appears in Appendix~\ref{app:conservatism_validation}.

\subsection{Hierarchical Inference and Groupings}

Evaluation data have a naturally nested structure: individual responses are nested within items (distinct prompts), which nest within groupings (model-task combinations). The framework operates simultaneously at both levels. At the item level, Bayesian inference characterises performance on each individual prompt across its repeated evaluations. At the grouping level, a hierarchical Bayesian model partially pools information across items, yielding stable estimates even as individual items terminate at different points.

Groupings are user-defined partitions of the evaluation space; each grouping maintains its own inference state and triggers stopping criteria independently. A factorial grouping design based on all manipulated factors that may systematically affect performance is generally recommended (e.g., model $\times$ task, or model $\times$ task $\times$ difficulty). Finer groupings provide higher resolution but require more observations to achieve precision thresholds; coarser groupings converge faster but may obscure performance heterogeneity.

\subsection{Inference Pathways}

The package supports three score types through tailored inference pathways, with distributional assumptions appropriate to the data type. Routing is determined by user configuration: binary inference is the default; ordinal inference is activated by declaring which tasks use ordered categorical scoring; and continuous inference is activated for tasks using score aggregation functions or continuous metrics. Table~\ref{tab:pathway_summary} summarises the three pathways.

\textbf{Binary} scores (correct/incorrect) are modelled with a logit-normal hierarchical structure at the grouping level, selected over the conjugate Beta-Binomial for computational reliability (see Appendix~\ref{app:model_comparison} for simulation evidence). The estimand is the population-level mean success probability.

\textbf{Ordinal} scores (discrete ordered categories, e.g., 0--10 rubrics) are modelled using a hierarchical ordered logistic (cumulative link) model, with a Dirichlet-Multinomial fallback when ordered assumptions are inappropriate. The estimand is the modal category of the item-averaged category distribution. Stopping employs a hybrid of two pathways: \textit{Pathway~1} requires both a narrow modal credible interval \emph{and} low distributional entropy, confirming a genuinely peaked distribution rather than a spurious narrow interval from sparse data; \textit{Pathway~2} evaluates whether the entropy credible interval width has fallen below a convergence threshold, handling distributions that are legitimately diffuse and lack a clear mode.

\textbf{Continuous bounded} scores (e.g., normalised similarity metrics, mean rubric aggregates) are modelled using a hierarchical logit-normal structure operating on item-level summary statistics rather than individual observations, providing approximately 20$\times$ theoretical computational speedup relative to observation-level inference.

\section{Empirical Validation}
\label{sec:validation}

To evaluate whether the stopping framework preserves statistical validity across inference pathways and the performance spectrum, we conducted a $3 \times 3$ matrix experiment crossing three inference pathways (binary, ordinal, and continuous; Table~\ref{tab:pathway_summary}) with three performance levels ($\hat{p}$:low, mid, and high). Each cell was run in \emph{shadow mode} - executing all 2{,}000 planned trials while internally tracking when stopping criteria would have been met. To isolate the pure truncation effect, we compare the score computed from the full run against the score computed from only those trials that would have been collected up to the stopping decision, eliminating between-run variance from LLM stochasticity.

The binary column employed distinct benchmarks and models to achieve natural performance variation: MATH Level~5 with GPT-3.5 Turbo (low, $\hat{p} \approx 0.05$), GPQA Diamond with GPT-4o (mid, $\hat{p} \approx 0.50$), and MMLU 0-shot with GPT-4o (high, $\hat{p} \approx 0.83$). The ordinal and continuous columns used WritingBench~\cite{wu2025writingbench} with Claude Sonnet 4.5\footnote{All models accessed via API in March 2026: Anthropic Claude Sonnet 4.5 (\texttt{claude-sonnet-4-5-20250929}), OpenAI GPT-4o (\texttt{gpt-4o-2024-08-06}), OpenAI GPT-3.5 Turbo (\texttt{gpt-3.5-turbo-0125}).}, varying \texttt{max\_tokens} (50, 500, 5{,}000) to modulate performance across an 11-category rubric ($K=11$, scores 0--10). All cells used 200 items (198 for mid-binary, the full GPQA Diamond set), 10 epochs per item, a precision threshold of $\delta = 0.05$, and 97\% credible intervals (see Appendix~\ref{app:matrix_design} for full experimental protocol).

\subsection{Stopping Efficiency and Score Fidelity}
\label{sec:efficiency_fidelity}

\begin{figure}[t]
\centering
\includegraphics[width=\columnwidth]{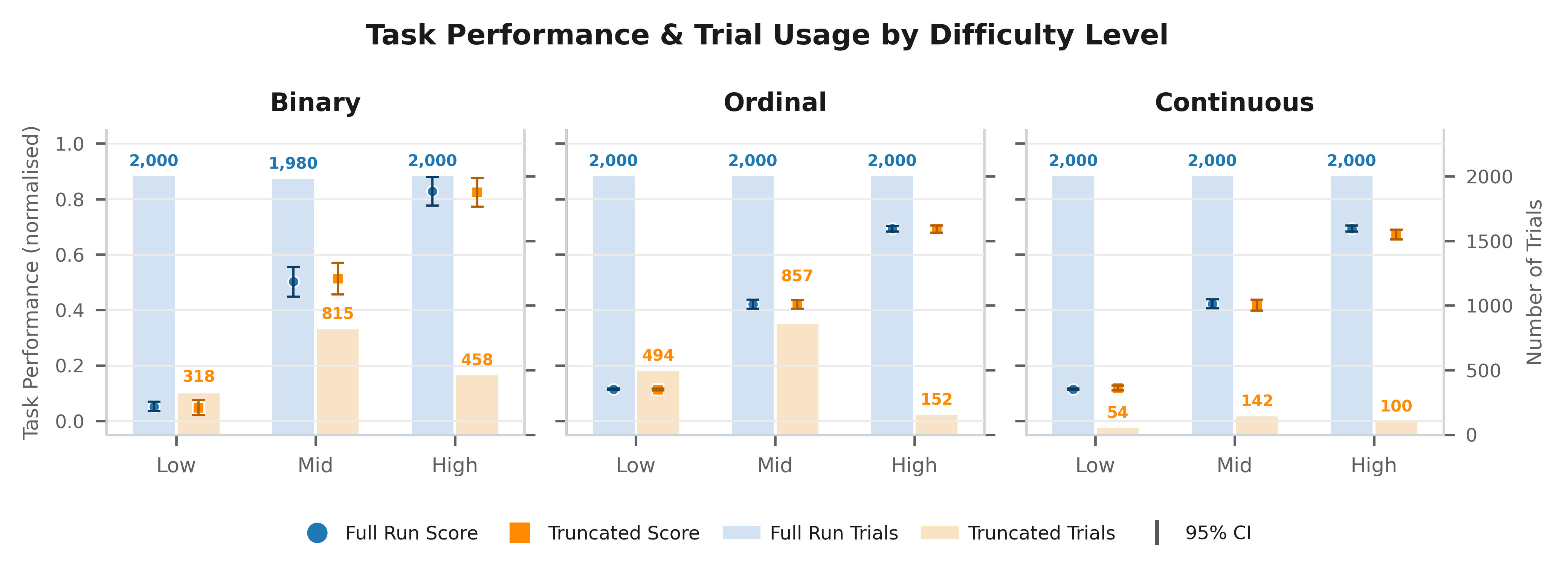}
\caption{Stopping efficiency and score fidelity across the $3 \times 3$ validation matrix. Bars show the number of trials: full run (blue, 2{,}000 planned; 1{,}980 for mid-binary) versus truncated (orange, with stop trial annotated). Overlaid markers show normalised task performance: full run (blue circles) versus truncated (orange squares). The close alignment of score markers demonstrates high fidelity despite substantial trial reductions, particularly in the continuous pathway (54--142 trials of 2{,}000). Error bars are item-level 95\% confidence intervals (mean $\pm$ 1.96\,SEM).}
\label{fig:performance_overlay}
\end{figure}

Figure~\ref{fig:performance_overlay} presents the core validation results across all nine cells. All nine cells triggered early stopping, with efficiency gains ranging from 57.2\% (mid-ordinal) to 97.3\% (low-continuous). Across all cells, the mean absolute score deviation between full-run and truncated estimates was 0.006 on a normalised $[0,1]$ scale, with a mean efficiency gain of 81.1\%.

Two patterns are evident. First, a \emph{pathway effect}: the continuous pathway achieves the highest efficiency (mean 95.1\%, range 93--97\%), followed by ordinal (75.0\%, 57--92\%) and binary (73.3\%, 59--84\%). The continuous pathway's advantage reflects the high information density of real-valued scores: each observation reduces posterior uncertainty substantially more than a binary outcome (at most 1~bit) or a discretised ordinal category. Second, an \emph{extremity effect}: within each pathway, cells with extreme performance levels (near floor or ceiling) yield greater efficiency than mid-range cells, because mid-range scores maximise per-item variance, slowing posterior convergence. Note that the asymmetric conservatism adjustment (Section~\ref{sec:conservatism}) was not exercised in this experiment: even the lowest-performing cell ($\hat{p} \approx 0.05$) sat above the 1\% low-performance threshold that engages it. Its behaviour is validated separately in Appendix~\ref{app:conservatism_validation}.

Despite these substantial efficiency gains, score fidelity remains high. The overlaid markers in Figure~\ref{fig:performance_overlay} show that full-run and truncated estimates are closely aligned across all cells. The ordinal pathway exhibits the smallest deviations (mean $|\Delta| = 0.001$), followed by binary (0.006) and continuous (0.011). Even the largest single-cell deviation (0.022, high-continuous) falls within the precision threshold ($\delta = 0.05$). The mid-ordinal cell is notable for triggering via Pathway~2 (entropy convergence) rather than Pathway~1 (modal CI), validating the dual-pathway design for distributions that lack a dominant mode (Appendix~\ref{app:ordinal_pathway_validation}). The deviations reported here are unpaired comparisons of each run's aggregate estimate, computed as the mean of per-item, epoch-averaged scores (the unit displayed in Figure~\ref{fig:performance_overlay}); the item-paired analysis in Section~\ref{sec:equivalence} isolates the per-item truncation effect.

\subsection{Statistical Equivalence}
\label{sec:equivalence}

To formally assess whether truncation introduces systematic bias, we applied Bayesian equivalence testing using the HDI (highest density interval) + ROPE (region of practical equivalence) framework~\cite{kruschke2018bayesian}. Per-cell analyses use 94\% HDIs following the HiBayES analysis convention; the pooled meta-analysis uses 97\% HDIs matching the \texttt{optstop} stopping criterion. For each cell, per-item epoch-averaged scores from the full run and the truncated run were paired by item identity, and a hierarchical meta-analysis pooled results across all nine cells to obtain a joint estimate of the overall truncation effect (see Appendix~\ref{app:analysis_a} for per-cell results and Appendix~\ref{app:hierarchical_meta} for full meta-analysis specification). We note there were two \textit{undecided} binary cells (low and mid), but this reflects the low information content of binary scoring (wide posteriors from at most 1~bit per observation), not truncation bias; there was an additional \textit{undecided} ordinal cell (high), which reflects a mode-vs-mean estimand mismatch. Full details can be found in Appendix~\ref{app:analysis_a}, and all three are resolved by the hierarchical meta-analysis.

\begin{figure}[t]
\centering
\includegraphics[width=\columnwidth]{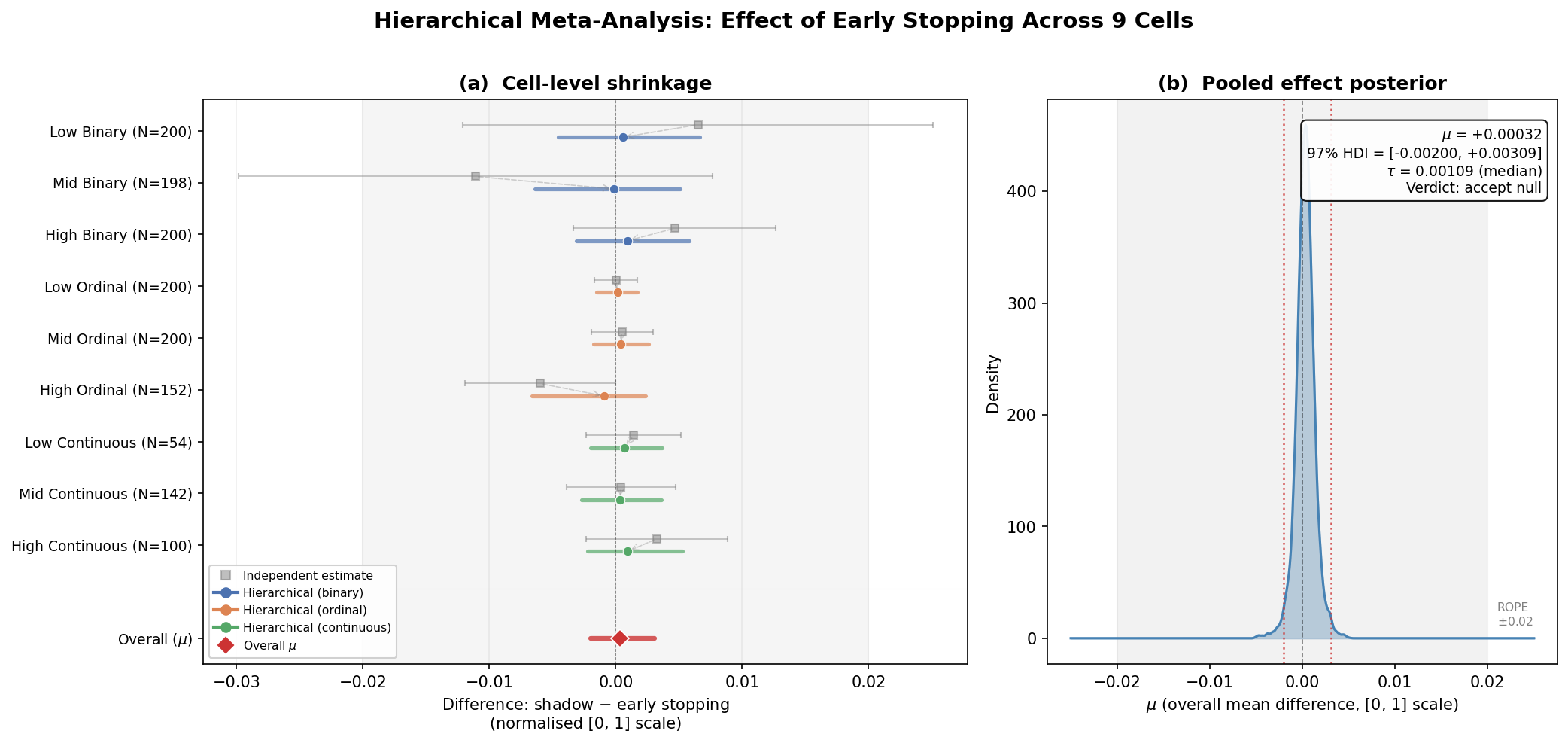}
\caption{Bayesian hierarchical meta-analysis of the truncation effect across all nine cells. (a)~Forest plot showing independent estimates (grey squares) and hierarchical estimates after partial pooling (coloured circles). Dashed arrows connect paired estimates, illustrating shrinkage toward the group mean. The bottom row shows the overall estimate $\mu$. Cell labels report $N$, the number of items matched in the paired comparison (fewer in cells that stopped before a full pass over all items). Grey error bars: Wald 95\% confidence intervals from observed standard errors; coloured error bars: 97\% HDIs from the hierarchical posterior. (b)~Posterior density of $\mu$ (overall mean difference, normalised $[0,1]$ scale). Shaded region: ROPE ($\pm 0.02$); vertical dashed lines: 97\% HDI boundaries. The entire HDI falls within the ROPE, yielding an accept-null verdict.}
\label{fig:hierarchical_meta}
\end{figure}

Figure~\ref{fig:hierarchical_meta} presents the results. Panel~(a) shows the forest plot with hierarchical shrinkage: grey squares indicate independent cell-level estimates (with Wald 95\% confidence intervals from observed standard errors), while coloured circles show the hierarchical estimates after partial pooling toward the group mean (with 97\% HDIs). The shrinkage is most pronounced for cells with larger standard errors - notably the low-binary, mid-binary, and high-ordinal cells, whose independent estimates are pulled toward the negligible group mean. Panel~(b) shows the posterior distribution of the overall mean truncation effect $\mu$. The estimate is $\hat{\mu} = {+}0.0003$ with a 97\% HDI of $[{-}0.002, {+}0.003]$ - entirely contained within the ROPE of $\pm 0.02$, yielding a clear accept-null verdict. The estimated between-cell heterogeneity is $\hat{\tau} = 0.001$, confirming consistency across cells.

As supplementary evidence, we verified that truncation preserves relative performance rankings - a property critical for capability comparisons. Using the ordinal and continuous columns, where the same benchmark (WritingBench) was evaluated at three \texttt{max\_tokens} levels, hierarchical models confirmed that the ranking $\texttt{max\_tokens} = 5{,}000 > 500 > 50$ was preserved with non-overlapping 94\% HDIs in both full-run and truncated conditions (Appendix~\ref{app:ranking}).

These results establish that the framework's stopping decisions preserve score fidelity across all cells and statistical equivalence at the pooled level, with the three individually undecided cells resolved by the hierarchical meta-analysis. The per-cell test is deliberately stringent - item-paired equivalence within a ROPE narrower than the framework's own precision target - and an undecided verdict indicates insufficient per-cell evidence rather than a detected difference: at a ROPE matching $\delta$, both binary cells accept equivalence (Appendix~\ref{app:rope_sensitivity}).

\section{Conclusions}
This paper has presented \texttt{optstop}, a precision-based optimal stopping framework for LLM evaluation that determines when data collection can safely terminate. Unlike adaptive item selection methods, the framework requires no pre-calibrated item banks and leaves every benchmark item eligible for evaluation, rather than pre-selecting a subset.

Empirical validation through a $3 \times 3$ matrix experiment (Section~\ref{sec:validation}) demonstrates that the framework can eliminate the majority of planned trials while preserving both score fidelity and statistical equivalence, with truncation bias confirmed as negligible by hierarchical meta-analysis. Efficiency gains are pathway-dependent, with continuous scores yielding the largest savings, followed by ordinal and then binary - reflecting the higher per-observation information content of real-valued and ordered-categorical data relative to single-bit binary outcomes. This pattern suggests that practitioners designing evaluations with efficiency in mind should prefer continuous or ordinal rubrics over binary pass/fail scoring where the assessment task permits it. These results were obtained under a configuration of 200 items and 10 epochs per item; the realised efficiency gains will vary with evaluation design, and leaner configurations (fewer items or epochs) will necessarily afford less scope for early termination while still requiring sufficient data for posterior convergence.

The framework's validity rests on exchangeability of observations within groupings and randomised presentation order. Both conditions can be satisfied in practice: the \texttt{inspect\_ai} framework provides a \texttt{sample\_shuffle} option for randomisation (not enabled by default; see Appendix~\ref{app:precision_stopping}), and apparent violations of exchangeability - for instance, systematic difficulty gradients within a task - are typically resolvable through finer grouping definitions that partition heterogeneous items into more homogeneous subsets (Appendix~\ref{app:groupings}). Credible interval coverage depends on the hierarchical model being well-specified, and near-boundary performance (close to 0\% or 100\%) may cause intervals to undercover due to shrinkage effects.

Presentation-order robustness testing (Appendix~\ref{app:order_robustness}) found 100\% empirical coverage across binary and continuous cells (90/90 permutation comparisons), consistent with the stopping criterion's design: because stopping fires when CI width reaches $\delta$, the resulting intervals are wide enough to absorb the small truncation biases observed (mean $|\Delta\theta| \leq 0.015$ across binary and continuous cells). This reflects the precision guarantee, not frequentist coverage calibration. The conservatism mechanism (Appendix~\ref{app:conservatism}) further extends data collection in the low-performance regime where undercoverage would be most acute (see validation of this mechanism in Appendix~\ref{app:conservatism_validation}), and the stopping criterion targets interval \emph{width} rather than coverage, so a narrow interval reflects posterior concentration even if shrinkage causes minor displacement from the true value. Indeed, controlled simulation (Appendix~\ref{app:model_comparison}) found empirical coverage of approximately 80\% at 94\% nominal for the hierarchical binary model under moderate heterogeneity - a gap that is symmetric across model choices and reflects finite-sample calibration rather than a model deficiency, but that reinforces the distinction between the framework's width guarantee and frequentist coverage.

The present validation, while spanning three inference pathways, three performance levels, and multiple benchmarks, represents a single experimental configuration - broader replication across evaluation contexts, scale sizes, and grouping structures would strengthen confidence in the framework's generalisability.

The \texttt{optstop} package is available as an open-source Python library. For evaluation campaigns where computational cost constrains the breadth or depth of assessment, the framework offers a principled mechanism to reclaim resources from well-characterised model-task combinations and redirect them toward regions of genuine uncertainty. Although developed for LLM evaluation, the approach extends to other domains with expensive, hierarchically structured data collection, provided the distributional assumptions are adapted. Knowing when to stop is not merely an optimisation - it is integral to evaluation regimes that are both rigorous and sustainable.

\paragraph{Code and data availability.} The \texttt{optstop} package, together with the simulation and validation code and the shadow-mode datasets underlying the figures and tables in this paper, is available at \url{https://github.com/UKGovernmentBEIS/optstop}.

\section*{Acknowledgements}

The author thanks Magda Dubois and Lennart Luettgau for valuable discussions and feedback throughout this work, JJ Allaire for building out \texttt{inspect\_ai} connectivity, and Cozmin Ududec for thoughtful comments on the manuscript.

\printbibliography[heading=bibintoc,title={References}]
\appendix

\section{Technical Methods: Stopping Framework}
\label{app:sec2_methods}

\subsection{Precision-based Stopping}
\label{app:precision_stopping}
The challenge of determining adequate sample sizes has occupied statisticians since the early twentieth century. Classical power analysis \cite{cohen1988edition} addresses this prospectively, calculating required samples to detect effects of specified size with desired probability. However, such approaches assume knowledge of effect sizes and variance structures that may be unavailable or inappropriate in the context of LLM evaluation, where performance characteristics vary dramatically across models, tasks, and prompt formulations.

An alternative tradition - sequential analysis - permits sample size determination during data collection rather than before it. Pioneered by Wald \cite{wald1945sequential} in the context of quality control, sequential methods evaluate accumulating evidence after each observation and terminate sampling when a decision criterion is satisfied. This approach offers substantial efficiency gains when the underlying signal is strong, as fewer observations are needed to reach confident conclusions. A key justification for adopting a Bayesian framework for sequential stopping is the Stopping Rule Principle \cite{barnard1947book,berger1988likelihood}: because the posterior depends only on the likelihood and prior, Bayesian inferences are valid regardless of the rule governing when data collection ceased.

Within sequential analysis, two broad families of stopping rules have emerged. Inference-based stopping rules terminate sampling when sufficient evidence exists to decide between competing hypotheses - for instance, when a Bayes factor exceeds a threshold \cite{schonbrodt2017sequential} or when a posterior probability crosses a decision boundary \cite{berry1985interim}. Such rules are well-suited to confirmatory contexts where the goal is to adjudicate between pre-specified alternatives.

Precision-based stopping rules, by contrast, terminate sampling when parameter estimates achieve desired accuracy, typically operationalised as confidence or credible interval width falling below a threshold \cite{kelley2006sample}. This approach aligns naturally with the exploratory and descriptive goals common in LLM evaluation, where the objective is often to characterise model performance with adequate precision rather than to test specific hypotheses. When an evaluator seeks to determine ``how well does this model perform on task X?'', the natural stopping point is when that performance estimate is sufficiently precise for the intended use - whether for model comparison, capability reporting, or safety assessment.

The \texttt{optstop} package implements precision-based stopping within a Bayesian inferential framework. Bayesian methods offer several advantages in this context: natural quantification of uncertainty through posterior distributions, coherent updating as data accumulate, and principled handling of small samples through prior regularisation. The core logic proceeds as follows: after each batch of observations, posterior distributions over performance parameters are computed via Markov Chain Monte Carlo sampling using PyMC~\cite{abril-pla2023pymc}, credible intervals are derived, and stopping criteria are evaluated. Sampling continues until precision thresholds are met or alternative termination conditions are satisfied.

Two critical assumptions underlie valid application of this framework to LLM evaluation. First, exchangeability within groupings: observations within a defined grouping (e.g., a specific model-task combination) should be exchangeable in the statistical sense - their joint distribution should be invariant to permutation. This assumption is generally reasonable when evaluation items are drawn from a well-defined population and model behaviour is consistent across the evaluation session. Second, randomised presentation order: the sequence in which items are evaluated should not systematically bias early versus late observations. The \texttt{inspect\_ai} framework supports this through its \texttt{sample\_shuffle} option (which accepts an optional seed for reproducibility), but shuffling is not enabled by default. This is particularly important when item IDs within a grouping are expected to systematically differ by incrementing ID (e.g., difficulty progression). Evaluators using \texttt{optstop} - whether through \texttt{inspect\_ai} or independently - should ensure randomised item ordering. Violations of these assumptions - for instance, through adaptive item selection or model drift during evaluation - may compromise the validity of stopping decisions.

\subsubsection{Credible Interval Widths}
\label{app:ci_widths}
The primary stopping criterion in \texttt{optstop} evaluates whether the width of the posterior credible interval for a performance parameter has fallen below a user-specified threshold. Formally, let $\theta$ denote the parameter of interest (e.g., success probability for a binary task), and let $[\theta_L, \theta_U]$ denote the $(1-\alpha)$ credible interval derived from the posterior distribution $p(\theta | \text{data})$. At the grouping level, this is the highest density interval (HDI); at the item level, equal-tailed quantile intervals are used for computational efficiency (see pathway-specific details in Appendix~\ref{app:pathways}). The credible interval width is simply:
$$W = \theta_U - \theta_L$$
Stopping occurs when $W < \delta$, where $\delta$ is the precision threshold specified by the user. The package exposes two such thresholds: \texttt{delta\_item} governs stopping at the individual item (sample) level (determining when sufficient epochs (repetitions) have been collected for a given item), while \texttt{delta\_cap} governs stopping at the grouping level (determining when sufficient items have been evaluated for a given model-task combination).
The interpretation of these thresholds should be calibrated to the evaluator's practical requirements. A threshold of $\delta = 0.05$ implies that the performance estimate is precise to within $\pm{2.5}$ percentage points at the specified credibility level (default 97\%, chosen to provide wider coverage than the frequentist 95\% convention while avoiding the efficiency cost of 99\% intervals). For high-stakes safety evaluations requiring fine discrimination between models, tighter thresholds may be appropriate; for preliminary capability surveys, looser thresholds may suffice. Note that the threshold represents a precision \emph{target} rather than an accuracy guarantee - the interval is narrow, but its location depends on the observed data. Furthermore, the nominal credibility level assumes the hierarchical model is well-specified; in practice, coverage depends on sample size and the proximity of true performance to boundaries. For smaller samples or near-boundary performance, hierarchical shrinkage may cause intervals to undercover relative to nominal levels (see Appendix~\ref{app:order_robustness} for empirical assessment).

The efficiency gains from this criterion depend on the underlying performance level and its variance. When performance is very high, posterior distributions concentrate rapidly and narrow credible intervals are achieved with fewer observations. When performance is very low, posteriors also concentrate rapidly, but this apparent precision may be misleading: rare successes may not yet have been observed, warranting the additional caution detailed in Appendix~\ref{app:conservatism}. When performance is moderate (near 50\% for binary outcomes) or highly variable, more observations are required. This behaviour is statistically appropriate: genuinely uncertain situations warrant more evidence, while clear-cut cases can be resolved quickly.

\subsubsection{Stabilisation Criterion}
\label{app:stabilisation}
The credible interval width criterion assumes that continued sampling will eventually yield a sufficiently narrow interval. While credible intervals necessarily converge given sufficient data, the convergence rate varies substantially with the underlying variability (as described above). In cases where convergence to the user's precision threshold would require impractically many observations - for instance, under high between-item heterogeneity or when the threshold is tight relative to the inherent posterior variance - additional sampling yields progressively diminishing precision gains.
To address such cases, \texttt{optstop} implements a secondary stopping criterion based on credible interval stabilisation. Rather than requiring the interval to be narrow in absolute terms, this criterion evaluates whether the interval width has ceased to decrease meaningfully with additional data - a signal that further sampling offers diminishing inferential returns.

The stabilisation criterion operates by tracking the trajectory of credible interval widths across successive inference updates. Let $W_1, W_2, \ldots, W_t$ denote the sequence of interval widths computed at each update. A linear regression is fitted to recent values within a sliding window of size $k$ (controlled by the \texttt{stab\_window} parameter), yielding a slope estimate $\hat{\beta}$ representing the rate of change in interval width:
$$\hat{\beta} = \frac{\sum_{i=1}^{k}(i - \bar{i})(W_{t-k+i} - \bar{W})}{\sum_{i=1}^{k}(i - \bar{i})^2}$$
Stabilisation is declared when $|\hat{\beta}| < \epsilon$, where $\epsilon$ is controlled by the \texttt{CI\_delta} parameter. Additionally, to guard against premature stabilisation declarations due to transient plateaus, the criterion requires that the slope trajectory itself has stabilised - specifically, that the slope is not trending toward steeper descent, which would indicate that precision gains are accelerating rather than diminishing. The stabilisation check also requires at least four accumulated slope estimates before evaluation, preventing noisy early estimates from triggering premature declarations.

This dual-criterion approach - absolute width or stabilised width - ensures that stopping decisions are appropriate across diverse performance distributions. Clear-cut cases with concentrated posteriors trigger the width criterion; ambiguous cases with irreducible uncertainty trigger the stabilisation criterion once further data collection becomes uninformative.

\subsubsection{Conservatism and pass@K}
\label{app:conservatism}
A systematic risk attends any early stopping procedure: premature termination may occur before rare but important events have been observed. In LLM evaluations, this risk manifests acutely in the assessment of difficult tasks where model success is infrequent. Consider evaluating a model's ability to solve challenging mathematical proofs, where the model might succeed on only 5\% of attempts. An early stopping rule optimising for precision might terminate sampling after observing a string of failures, concluding with high confidence that the success rate is near zero - potentially missing the model's genuine (if limited) capability.
This concern connects to the pass@k evaluation paradigm \cite{chen2021evaluating}, where model capability is assessed by whether at least one success occurs across $k$ independent attempts. Under pass@k, rare successes carry substantial inferential weight: a model that succeeds once in twenty attempts demonstrates qualitatively different capability than one that never succeeds. Early stopping procedures must therefore exercise particular caution when observed performance is low, as the most decision-relevant observations (rare successes) are precisely those least likely to have occurred in limited samples.
The \texttt{optstop} package addresses the most acute form of this risk - when estimated performance is at or near zero, suggesting the model may have no genuine capability on the task - through an asymmetric conservatism adjustment applied when estimated performance falls below a threshold (controlled by \texttt{low\_performance\_threshold}, defaulting to 1\%). When this condition is met, stopping criteria are modified to require stronger evidence before termination:
\begin{enumerate}
    \item Inflated effective interval width: The computed credible interval width is multiplied by a conservatism factor $c > 1$ (controlled by the \texttt{conservatism} parameter, defaulting to 5) before comparison against the stopping threshold. This inflation means that nominally narrow intervals no longer satisfy the stopping criterion, requiring additional data collection.
    \item Tightened stabilisation threshold: The slope threshold for the stabilisation criterion is divided by the conservatism factor ($|\hat{\beta}| < \epsilon / c$), requiring more convincing evidence of plateau before stabilisation-based stopping is permitted.
\end{enumerate}

The effect of these adjustments is to delay stopping decisions for low-performing model-task combinations, providing additional opportunity for rare successes to be observed. The conservatism adjustment is asymmetric: it applies only when performance is poor, not when performance is high. A model that succeeds on 99\% of attempts can be confidently characterised with fewer observations than one that succeeds on 1\% of attempts, reflecting the differential inferential demands of these situations.

This asymmetry aligns with the pragmatics of capability evaluation. When assessing whether a model can perform a task (capability detection), rare successes are highly informative; when assessing how well a model performs a task it clearly can do (capability measurement), rare failures are less decision-relevant. The conservatism mechanism encodes this asymmetry directly into the stopping rules. Note that \texttt{conservatism} and \texttt{low\_performance\_threshold} are coupled parameters: for a given data budget, the effective CI-width target ($\delta / c$) must remain achievable. With defaults ($c = 5$, $\delta = 0.05$), the effective target is 0.01 - achievable with moderate data. Appendix~\ref{app:conservatism_validation} presents a dedicated sensitivity analysis isolating the conservatism mechanism across calibrated performance levels.

\subsection{Hierarchical Inference}
\label{app:hierarchical_inference}
LLM evaluation data possess a natural hierarchical structure: individual responses (observations) are nested within items (distinct prompts or problems), which are themselves nested within evaluation groupings (model-task combinations). This structure motivates a hierarchical approach to inference and stopping decisions, operating simultaneously at multiple levels of aggregation.

At the item level, the unit of analysis is a single evaluation item (e.g., one mathematical problem or one coding challenge) assessed across multiple epochs - repeated evaluations of the same item, potentially with varied sampling parameters or random seeds. The inferential goal at this level is to characterise model performance on that specific item with adequate precision. When the credible interval for an item's success rate (or score distribution) becomes sufficiently narrow, further epochs for that item are unnecessary - the model's behaviour on that particular input is well-characterised.

At the grouping level, the unit of analysis is an aggregation of items sharing common characteristics - typically a specific model evaluated on a specific task, though more granular groupings (e.g., by difficulty level or topic) are supported. The inferential goal at this level is to characterise the model's overall performance on the task, integrating information across all constituent items. When the credible interval for the grouping's aggregate performance becomes sufficiently narrow, the evaluation of that model-task combination can terminate entirely, with remaining (unevaluated) items skipped.

This hierarchical structure yields a natural cascade of stopping decisions. Early in an evaluation run, item-level stopping criteria begin to trigger for individual items where model behaviour is consistent across epochs. As item-level data accumulates, grouping-level inference integrates these observations, and grouping-level stopping criteria may trigger for model-task combinations where aggregate performance is well-characterised. The evaluation thus progressively focuses computational resources on the most uncertain regions of the evaluation space: items with variable behaviour and groupings with heterogeneous performance across items.

The \texttt{optstop} package implements this hierarchy through distinct but coordinated inference processes at each level. Item-level inference operates on the epoch-wise observations for a single item, updating a posterior distribution over that item's performance parameter after each epoch. Grouping-level inference operates on summary statistics aggregated across items, employing hierarchical Bayesian models that partially pool information across items while respecting their individual characteristics.

The hierarchical Bayesian approach offers particular advantages for grouping-level inference. Consider estimating an LLM's overall success probability on a task comprising 100 distinct items. A naive approach might simply pool all observations; treating successes and failures as exchangeable draws from a single Bernoulli distribution. However, this ignores meaningful variation across items: some problems are inherently easier than others, and model performance varies accordingly. At the opposite extreme, treating each item's success probability as entirely independent discards the commonality that they all represent the same model's capability on related problems.

Hierarchical models navigate between these extremes through partial pooling. Item-level parameters are modelled as draws from a population distribution whose parameters are themselves estimated from the data. Items with limited observations are ``shrunk'' toward the population mean, borrowing strength from other items; items with extensive observations retain estimates closer to their individual data. This approach yields more stable grouping-level estimates than either complete pooling or no pooling, particularly when item sample sizes are uneven - as they inevitably become under adaptive stopping, where easy items terminate quickly and difficult items accumulate more observations. For example, a mathematical reasoning task might include both routine calculations (high success probability) and complex proofs (low success probability); hierarchical inference appropriately weights these when characterising overall task performance.

\subsection{Leveraging Groupings}
\label{app:groupings}
The effectiveness of any optimal stopping procedure depends critically on the definition of the units across which stopping decisions operate. In LLM evaluation, this choice is far from trivial: the ``jagged frontier'' of model capabilities \cite{dell2023navigating} - a metaphor borrowed from the study of AI-augmented professional tasks - means that performance varies dramatically across tasks, domains, prompt formulations, and even superficial features of evaluation items. A model that excels at arithmetic may struggle with algebra; one that handles formal English may falter with colloquial text. Aggregating across these dimensions risks masking important heterogeneity, while disaggregating too finely may yield sample sizes too small for reliable inference.

The \texttt{optstop} package addresses this through flexible, user-defined groupings - partitions of the evaluation space within which stopping decisions operate independently. Each grouping maintains its own inference state, accumulates its own observations, and triggers stopping criteria according to its own precision trajectory. When a grouping's credible interval achieves the required precision, evaluation of that grouping terminates while other groupings continue.

Groupings can be defined along any dimension captured in the evaluation metadata. A general rule of thumb is to use a factorial design based on all manipulated factors that may systematically impact performance. Common configurations include (but are not limited to):

\begin{itemize}
    \item Model $\times$ Task: Each model-task combination constitutes a separate grouping, appropriate when the primary goal is to characterise each model's performance on each task independently.
    \item Model $\times$ Task $\times$ Difficulty: Further stratification by difficulty level or sub-task category, useful when capability variation within tasks is substantial.
    \item Tag-based: Groupings defined by evaluation tags (e.g., ``safety-critical'', ``reasoning'', ``factual-recall''), enabling domain-specific stopping thresholds and conservatism settings.
\end{itemize}

The choice of grouping structure involves trade-offs between inferential resolution and statistical power. Finer groupings provide more granular capability characterisation but require more observations per grouping to achieve precision thresholds; coarser groupings achieve precision more quickly but may obscure important heterogeneity. These trade-offs interact with the conservatism considerations discussed in Appendix~\ref{app:conservatism}: groupings exhibiting low performance will accumulate additional observations under the conservatism adjustment, naturally directing computational resources toward capability boundaries where uncertainty is highest.

The grouping structure determines not only where stopping decisions are made but also what those decisions mean. A stopping decision for a fine-grained grouping (e.g., ``GPT-4 on multi-step arithmetic with chain-of-thought prompting'') provides a precise capability statement about a specific configuration. A stopping decision for a coarse grouping (e.g., ``GPT-4 on mathematical reasoning'') provides a broader but potentially less actionable summary. Evaluators should select grouping structures that align with the decisions their evaluation is intended to inform.
The independence of grouping-level inference also provides natural parallelism in the evaluation process. As different groupings reach stopping criteria at different times, the evaluation scheduler in \texttt{inspect\_ai} implementations can redistribute computational resources to remaining groupings. The evaluation thus progressively concentrates effort on the most uncertain regions of the capability space, where additional observations provide the greatest inferential value.

\subsection{Inference Pathways}
\label{app:pathways}
Evaluation scores in LLM assessment take diverse forms. Some tasks yield binary outcomes - the model either produces the correct answer or it does not (or is classified as such via a threshold). Others employ rubric-based scoring, where human or model judges assign discrete ratings (e.g., 1--5 quality scores or 0--10 capability assessments). Still others produce continuous metrics, such as BLEU scores, embedding similarities, or calibrated probability estimates. A general-purpose stopping framework must accommodate this diversity, routing each score type to appropriate inferential machinery while presenting a unified interface to the evaluator.

The package implements inference pathway routing based on user configuration and score context. Routing is determined by user-specified parameters - \texttt{ordinal\_tasks} substring patterns for ordinal scored tasks, and for continuous tasks, \texttt{continuous\_tasks} substring patterns (standalone mode), and the \texttt{score\_agg} aggregation setting (via \texttt{inspect\_ai} integration) - with binary inference as the default. This routing proceeds according to the following rules:

\begin{itemize}
    \item Binary pathway: The default, engaged when no ordinal or continuous configuration matches the grouping.
    \item Ordinal pathway: Engaged when the grouping name matches a user-provided \texttt{ordinal\_tasks} substring pattern and scores are not aggregated.
    \item Bounded continuous pathway: Engaged when a score aggregation function (mean, median) is applied (\texttt{score\_agg}, via the \texttt{inspect\_ai} integration), or when the grouping name matches a \texttt{continuous\_tasks} substring pattern (standalone mode).
\end{itemize}
 
The routing decision is made at the grouping level and persists throughout the evaluation. Because routing depends on configuration parameters rather than observed score values, users with non-binary scoring schemes must declare these; otherwise all groupings default to binary inference regardless of the actual score distribution.

Each pathway implements the same conceptual framework - Bayesian inference yielding credible intervals evaluated against precision thresholds - but with distributional assumptions and model structures appropriate to the data type. The following subsections detail these pathway-specific implementations.

\subsubsection{Binary}
\label{app:binary_pathway}
Binary scoring represents the most common evaluation paradigm: the model's response is judged correct (1) or incorrect (0), with no intermediate gradations. Tasks employing exact-match evaluation, binary classifiers, or pass/fail rubrics generate binary scores. The inferential goal is to estimate the underlying success probability $\theta \in [0, 1]$ with adequate precision.

\paragraph{Identification.} The binary pathway is the default inference mode. Score type is determined once per grouping at initialisation and does not change during the evaluation.

\paragraph{Item-Level Inference.} For a single evaluation item assessed across multiple epochs, let $s$ denote the number of successes (score = 1) and $n$ the total number of trials observed. The \texttt{optstop} package employs an adaptive data-dependent prior where the prior distribution adjusts based on accumulating within-item evidence.

The prior parameters are set proportional to the current point estimate $\hat{p} = s/n$, with influence that decays exponentially as sample size increases:
$$\alpha_{\text{prior}} = \max\left(\gamma \cdot \hat{p},  0.5\right), \quad \beta_{\text{prior}} = \max\left(\gamma \cdot (1 - \hat{p}),  0.5\right)$$
where $\gamma = b \cdot \exp(-n / 10)$ is a scaling factor that diminishes with sample size, and $b$ is a base strength parameter (default 2). The floor of 0.5 corresponds to the Jeffreys non-informative prior, preventing the adaptive component from producing weaker regularisation than this baseline. This formulation ensures that the prior exerts meaningful regularisation with small samples but becomes increasingly dominated by likelihood as evidence accumulates - a form of vanishing prior influence. Note that because the prior parameters depend on the observed data, this is not a prior in the strict Bayesian sense; the construction is better understood as an empirical Bayes regularisation device. The Stopping Rule Principle (Appendix~\ref{app:precision_stopping}) strictly applies to models with fixed priors - a condition satisfied by the grouping-level hierarchical models (Appendix~\ref{app:binary_pathway}, Grouping-Level Inference) but not by this data-dependent item-level construction. The item-level procedure is better justified by the vanishing influence of the adaptive prior: because $\gamma$ decays exponentially with $n$, the posterior is increasingly determined by the likelihood alone, and by the time an item's CI is narrow enough to trigger stopping, the prior's contribution is negligible. The consequential stopping decisions - at the grouping level - rest on standard hierarchical posteriors to which the SRP applies without qualification.
The posterior distribution is then:
$$\theta_{\text{item}} \sim \text{Beta}(\alpha_{\text{prior}} + s,  \beta_{\text{prior}} + (n - s))$$
Credible intervals are computed via Monte Carlo sampling from this posterior, with equal-tailed quantiles extracted at the desired credibility level (in contrast to the highest density intervals used at the grouping level; see Appendix~\ref{app:ci_widths}). When conservatism is active (estimated performance below the \texttt{low\_performance\_threshold}; see Appendix~\ref{app:conservatism}), the prior is modified: the decay rate slows to $\gamma = b \cdot \exp(-n / (10c))$ where $c > 1$ is the \texttt{conservatism} parameter (default $c = 5$), and the data-dependent component of $\alpha_{\text{prior}}$ is multiplied by $c$ before the floor is applied. Under default parameterisation, this multiplicative boost is absorbed by the $\max(\cdot, 0.5)$ floor when $\hat{p}$ is very small, so the dominant item-level conservatism effects are the slowed prior decay and the CI width inflation: the effective CI width is multiplied by $c$ before comparison against stopping thresholds.

\paragraph{Grouping-Level Inference.} At the grouping level, observations are aggregated across items using a hierarchical model with logit-normal structure. Let $i = 1, \ldots, N$ index the items within the grouping, with $s_i$ successes observed across $n_i$ epochs for item $i$. The hierarchical model takes the form:
$$\mu_{\text{group}} \sim \text{Normal}(\mu_0, \sigma_0)$$
$$\sigma_{\text{group}} \sim \text{Exponential}(\lambda)$$
$$z_i \sim \text{Normal}(0, 1)$$
$$\eta_i = \mu_{\text{group}} + \sigma_{\text{group}} \cdot z_i$$
$$\theta_i = \text{logit}^{-1}(\text{clip}(\eta_i, -6, 6)) \quad \text{(numerical stability safeguard, bounding } \theta_i \in [0.0025, 0.9975]\text{)}$$
$$s_i \sim \text{Binomial}(n_i, \theta_i)$$
where the default prior hyperparameters are $\mu_0 = 0$, $\sigma_0 = 1.5$, and $\lambda = 1.0$. The zero-centred prior mean on the logit scale ($\mu_0 = 0$, corresponding to 50\% prior expected success rate) provides a weakly informative default that does not favour high or low performance, placing approximately 95\% of prior mass on success probabilities between 0.05 and 0.95. Users may adjust this via the \texttt{prior\_mu} parameter: when re-evaluating a model on a benchmark for which previous results are available, setting $\mu_0$ to the logit of the known performance level (e.g., $\text{logit}(0.75) \approx 1.1$) improves the accuracy of early point estimates and credible interval placement by centring the posterior near the true value from the outset. Because credible intervals are computed on the probability scale via the logistic transform, a well-placed posterior also yields narrower intervals than one centred near 50\% at the same precision level, which can lead to earlier stopping - though the magnitude of this effect depends on how far true performance is from the uninformed mid-point, and how well the prior maps onto it. The default prior width ($\sigma_0 = 1.5$; adjustable via \texttt{prior\_sigma}) ensures that a mis-specified prior is overridden by data within a modest number of items (of order 10--20 under moderate performance, though the exact rate depends on between-item heterogeneity and epochs per item), so the cost of an incorrect assumption is bounded. This parameter governs only the grouping-level hierarchical prior; the item-level adaptive prior (above) is unaffected. 

Further, $\mu_{\text{group}}$ represents the population-level mean on the logit scale, and $\sigma_{\text{group}}$ governs between-item heterogeneity. The non-centred parameterisation (introducing auxiliary variables $z_i$ rather than sampling $\eta_i$ directly) facilitates efficient MCMC sampling, particularly when between-item variance is small. The item-level success probabilities $\theta_i$ are obtained by applying the inverse logit (sigmoid) transformation to the latent linear predictor $\eta_i$.

The logit-normal hierarchical structure was selected over the conjugate Beta-Binomial alternative following controlled simulation comparison (see Appendix~\ref{app:model_comparison}). Under a Beta-Binomial data generating process (inherently favouring that model), the two approaches proved statistically indistinguishable in bias, coverage, and CI width across the mid-range (0.1--0.9 true performance; bias ratio 1.00, CI width ratio 1.00, coverage 0.80 vs.\ 0.81). Near boundaries (0.05, 0.95) the models remained closely matched. At exact boundaries (0.0 or 1.0) - where all items share identical success probability and between-item heterogeneity is zero - both models exhibit zero nominal coverage, reflecting fundamental information limitations when sparse binary data cannot distinguish true homogeneity from sampling coincidence. The Beta-Binomial exhibited 120--9{,}206$\times$ more MCMC divergences than the logit-normal across conditions, with effective sample sizes as low as 7 (vs consistently above 3{,}700 for the logit-normal), indicating substantially worse posterior geometry. The logit-normal therefore provides equivalent estimation quality with dramatically better computational reliability - the appropriate default for an adaptive framework where MCMC must run reliably across diverse and unknown data regimes.

The population-level success probability (the primary target for grouping-level inference) is the expected group accuracy: $\Theta = \frac{1}{N}\sum_{i=1}^{N} \theta_i$, computed as the mean of item-level success probabilities across posterior draws. This correctly accounts for between-item heterogeneity ($\sigma_{\text{group}}$); the simpler transformation $\text{logit}^{-1}(\mu_{\text{group}})$, which represents the success probability of a typical item ($z_i = 0$), can diverge from $\Theta$ when heterogeneity is large (by Jensen's inequality). Posterior inference proceeds via Markov Chain Monte Carlo sampling, yielding draws from $p(\Theta \mid s_i, n_i)$. The credible interval for $\Theta$ provides the basis for grouping-level stopping decisions.

\paragraph{Stopping Criteria.} Both the CI width criterion and the stabilisation criterion (Appendix~\ref{app:ci_widths}--\ref{app:stabilisation}) apply at item and grouping levels. At the item level, Monte Carlo sampling from the adaptive Beta posterior permits rapid CI computation without full MCMC machinery. At the grouping level, MCMC sampling is triggered at configurable intervals to update the hierarchical posterior and evaluate stopping conditions. The CI width inflation and slope strictness under conservatism (Appendix~\ref{app:conservatism}) applies at the grouping level, multiplying the MCMC-derived CI width by $c$ before comparison against \texttt{delta\_cap}, and dividing the slope threshold \texttt{CI\_delta} by $c$, accordingly.

\subsubsection{Ordinal (Discrete)}
\label{app:ordinal_pathway}
Many evaluation rubrics employ discrete ordered categories - quality ratings from 1--5, capability scores from 0--10, or multi-level correctness assessments (incorrect / partially correct / fully correct). Such ordinal scores occupy a middle ground between binary and continuous data: they preserve rank ordering (a score of 4 indicates better performance than a score of 3) but do not necessarily imply equal intervals (the difference between 3 and 4 may not equal the difference between 4 and 5 in any meaningful sense).

This ordinal structure presents both opportunities and challenges for inference. Treating ordinal scores as continuous (e.g., computing means) imposes interval-scale assumptions that may be unwarranted. Treating them as unordered categories (multinomial models) discards the rank information. Appropriate inference requires models that respect ordinality without assuming cardinality - a class of models well-developed in psychometrics and survey research \cite{agresti2010analysis}.

\paragraph{Identification.} The ordinal pathway is selected when the grouping name matches a user-provided \texttt{ordinal\_tasks} substring pattern (case-insensitive) and scores are not aggregated. The number of modelled categories is $K = \texttt{ordinal\_max\_score} + 1$ (default 11 for a 0--10 scale); \texttt{ordinal\_max\_score} should reflect the maximum possible score for the rubric in use.

\paragraph{Item-Level Inference.} Item-level inference for ordinal scores focuses on characterising the modal category - the most probable score value - and the uncertainty surrounding this characterisation. The inferential target is the integer category that represents typical performance, with the credible interval quantifying uncertainty about which category this is.

For a single item with observed scores $y_1, \ldots, y_n$ across $n$ epochs, a Bayesian bootstrap procedure \cite{rubin1981bayesian} estimates the posterior distribution over the modal category. The procedure operates as follows:

\begin{enumerate}
    \item Draw Dirichlet-distributed weights over observations: $\mathbf{w} \sim \text{Dirichlet}(\mathbf{1}_n)$
    \item Construct a weighted histogram across categories using these weights
    \item Identify the modal category as $\hat{m} = \arg\max_k (\text{weighted count in category } k)$
    \item Repeat for $B = 10{,}000$ bootstrap samples of the modal category
\end{enumerate}

The resulting distribution over modal categories $\hat{m}^{(1)}, \ldots, \hat{m}^{(B)}$ is summarised via percentile credible intervals. This CI is on the category scale - a CI of [6, 8] on a 0--10 scale indicates 97\% posterior probability (at the default credibility level) that the modal category lies between 6 and 8. For comparison against precision thresholds, the interval is normalised to $[0, 1]$ by dividing by the maximum score value $(K - 1)$.
A sample-size-scaled floor prevents premature stopping when bootstrap variance is zero due to homogeneous data: $W_{\min} = 1/((K-1) \cdot \sqrt{n})$. This floor decreases with sample size, reflecting that more agreeing observations genuinely warrant greater confidence.

\paragraph{Grouping-Level Inference.} Grouping-level inference for ordinal data employs hierarchical models that partially pool information across items while respecting ordinal structure. The \texttt{optstop} package implements two model variants:

\textit{Ordered Logistic (Cumulative Link) Model}: This model, derived from the psychometric literature \cite{mccullagh1980regression}, posits a latent continuous ability scale underlying the observed ordinal responses. The hierarchical structure places a population-level distribution over item abilities using non-centred parameterisation:
$$\mu_{\text{group}} \sim \text{Normal}(\mu_0, 2)$$
$$\sigma_{\text{group}} \sim \text{Exponential}(\text{rate}=1)$$
$$z_i \sim \text{Normal}(0, 1)$$
$$\eta_i = \mu_{\text{group}} + \sigma_{\text{group}} \cdot z_i$$
The default prior standard deviation of 2 on the group mean ($\sigma_0 = 2$; adjustable via \texttt{prior\_sigma}) is wider than for binary models ($\sigma_0 = 1.5$), reflecting the broader latent scale for ordinal responses. The default $\mu_0 = 0$ is weakly informative; users may adjust both parameters via \texttt{prior\_mu} and \texttt{prior\_sigma} (see Appendix~\ref{app:binary_pathway} for informed prior guidance). For each latent ability $\eta_i$, the probability of observing category $k$ or lower is given by a cumulative logistic function:
$$P(Y_i \leq k) = \text{logit}^{-1}(c_k - \eta_i), \quad k = 0, \ldots, K-2$$
where $c_0 < c_1 < \cdots < c_{K-2}$ are $K-1$ ordered cutpoints partitioning the latent scale into $K$ categories. To ensure identifiability, the first cutpoint is fixed at zero ($c_0 = 0$), with subsequent cutpoints parameterised as cumulative sums of positive increments: $c_k = \sum_{j=1}^{k} \gamma_j$ for $k \geq 1$, where $\gamma_j = \text{softplus}(\gamma_j^{\text{raw}})$ ensures positivity ($K-2$ free increments). The raw increments $\gamma_j^{\text{raw}}$ are given Normal priors whose mean and variance scale adaptively with $K$ to ensure reasonable cutpoint spacing across different rubric sizes. The category probabilities are then:
$$P(Y_i = k) = P(Y_i \leq k) - P(Y_i \leq k-1)$$
with boundary conditions $P(Y_i \leq -1) = 0$ and $P(Y_i \leq K-1) = 1$.
Given aggregated category counts $\mathbf{f}_i = (f_{i,0}, \ldots, f_{i,K-1})$ for each item $i$, the likelihood is multinomial:
$$\mathbf{f}_i \sim \text{Multinomial}(n_i, \mathbf{p}_i)$$
where $\mathbf{p}_i$ is the vector of category probabilities derived from the cumulative model. The population-level estimand is the modal category of the item-averaged probability vector: $\hat{m}_{\text{group}} = \arg\max_k \bar{p}_k$, where $\bar{p}_k = (1/N)\sum_{i=1}^{N} p_{i,k}$.

\textit{Dirichlet-Multinomial Model}: An alternative model treats ordinal categories as exchangeable (ignoring rank structure) but introduces hierarchical pooling across items:
$$\boldsymbol{\alpha}_{\text{group}} \sim \text{Dirichlet}(\mathbf{1}_{K})$$
$$\kappa \sim \text{Gamma}(\text{shape}=2,\; \text{rate}=0.1) \quad (\mathbb{E}[\kappa] = 20)$$
$$\mathbf{p}_i \sim \text{Dirichlet}(\kappa \cdot \boldsymbol{\alpha}_{\text{group}})$$
$$\mathbf{f}_i \sim \text{Multinomial}(n_i, \mathbf{p}_i)$$
This model is available as an alternative when ordered logistic assumptions are inappropriate (e.g., strongly bimodal response distributions that violate the latent-continuum assumption) or when MCMC sampling for the ordered logistic model encounters convergence difficulties. The population-level estimand for the Dirichlet model is $\hat{m}_{\text{group}} = \arg\max_k \alpha_{\text{group},k}$, corresponding to the mode of the expected category distribution.

Note that in both such cases, the appropriate estimand for stopping decisions may not correspond to the evaluator's intended performance estimand (i.e., whilst modal categories are appropriate for stopping decisions, the evaluator may still wish to report/use the mean across categories). This is not a problem (the evaluator can still calculate and use the mean from the stopped data), but the two should not be confused in reporting.

\paragraph{Stopping Criteria (Hybrid Approach).} Ordinal inference employs a hybrid stopping criterion addressing the distinctive challenges of ordered categorical data. Two pathways to stopping are evaluated:

\textit{Pathway~1 (Modal CI with Entropy Validation)}: The primary criterion evaluates modal credible interval width. However, a narrow modal CI can arise spuriously when limited data happen to concentrate in one category, even though the true distribution may be diffuse. To guard against such ``false peaks,'' Pathway~1 includes an entropy validation gate:

\begin{itemize}
    \item Compute Shannon entropy of the estimated category distribution: $H = -\sum_k p_k \log_2 p_k$ (in bits)
    \item If modal CI width is below threshold BUT the posterior median entropy exceeds a peakedness threshold (default 80\% of $\log_2 K$ bits), the narrow CI is deemed unreliable and stopping is not triggered
    \item Stopping via Pathway~1 requires both narrow modal CI AND low entropy, confirming a genuinely peaked distribution
\end{itemize}
  
\textit{Pathway~2 (Entropy Convergence)}: For distributions that are legitimately diffuse - spread across multiple categories without a clear mode - the modal CI may never achieve narrow thresholds. Pathway~2 evaluates whether the entropy credible interval width (on the $[0, 1]$ normalised entropy scale) has fallen below a convergence threshold (default 0.10, corresponding to $\pm5\%$ precision on the entropy scale). This requires at least three inference updates to guard against unreliable early MCMC estimates. Unlike rate-of-change criteria, which may plateau before convergence under exponential posterior contraction, this absolute-width criterion is satisfied in finite time under standard posterior concentration, provided the model is well-specified and data are sufficiently informative.

This dual-pathway approach ensures appropriate stopping across diverse distributional shapes: peaked distributions trigger Pathway~1; diffuse but converged distributions trigger Pathway~2; distributions still in flux (entropy CI still wide) continue data collection.

\subsubsection{Bounded Continuous}
\label{app:continuous_pathway}
Some evaluation metrics yield continuous values within known bounds - embedding cosine similarities (after rescaling to $[0, 1]$), normalised BLEU scores in $[0, 1]$, or calibrated probability estimates. Additionally, when evaluators request aggregation of discrete scores across sub-components (e.g., mean rubric scores across multiple criteria), the resulting averages are effectively continuous on a bounded interval.

\paragraph{Identification.} The bounded continuous pathway is selected when a score aggregation function (\texttt{score\_agg}: mean or median) is applied via the \texttt{inspect\_ai} integration, converting discrete sub-scores to continuous aggregates, or when the grouping name matches a \texttt{continuous\_tasks} substring pattern in standalone mode. Scores are normalised to the unit interval $[0, 1]$ for inference, with results transformed back to the original scale for reporting.

\paragraph{Item-Level Inference.} For continuous bounded scores, item-level inference employs a Beta distribution model after normalisation to $[0, 1]$. Let $y_1, \ldots, y_n$ denote the $n$ observed scores for an item, normalised to the unit interval. The sample mean $\bar{y}$ and variance $s^2$ inform Beta posterior parameters via method-of-moments estimation:
Given sample statistics, the posterior is approximated as:
$$\theta_{\text{item}} \sim \text{Beta}(\alpha_{\text{post}}, \beta_{\text{post}})$$
where $\alpha_{\text{post}}$ and $\beta_{\text{post}}$ are derived from moment-matching to the observed data, incorporating prior regularisation that decays with sample size (analogous to the binary case). Credible intervals are computed via Monte Carlo sampling from this posterior. The CI width, returned in the original score scale, is normalised to $[0, 1]$ by dividing by the score range $(U - L)$ before comparison against \texttt{delta\_item}.

\paragraph{Grouping-Level Inference.} Grouping-level inference for continuous bounded scores employs a hierarchical model that aggregates information across items while dramatically reducing computational cost (relative to ordinal equivalents). Rather than modelling individual observations (which could number in the thousands), the model operates on item-level summary statistics - specifically, the mean score for each item.

The hierarchical structure uses logit-normal parameterisation with non-centred form:
$$\mu_{\text{group}} \sim \text{Normal}(\mu_0, \sigma_0)$$
$$\sigma_{\text{group}} \sim \text{Exponential}(\lambda_\sigma)$$
$$\phi_{\text{group}} \sim \text{Gamma}(\text{shape}=\alpha_\phi,\; \text{rate}=\beta_\phi)$$
$$z_i \sim \text{Normal}(0, 1)$$
$$\mu_i = \text{logit}^{-1}(\text{clip}(\mu_{\text{group}} + \sigma_{\text{group}} \cdot z_i,\; -6, 6))$$
where the default prior hyperparameters are $\sigma_0 = 1.5$, $\lambda_\sigma = 1.0$, $\alpha_\phi = 2$, and $\beta_\phi = 1.0$. The default prior mean $\mu_0 = 0$ (corresponding to 50\% on the probability scale) reflects a neutral prior expectation for continuous scores; both $\mu_0$ and $\sigma_0$ are configurable via \texttt{prior\_mu} and \texttt{prior\_sigma} (see Appendix~\ref{app:binary_pathway}). Here, $\mu_{\text{group}}$ is the population mean on the logit scale, $\sigma_{\text{group}}$ governs between-item heterogeneity, and $\phi_{\text{group}}$ is a precision (concentration) parameter controlling within-item variability. The item-level means $\mu_i$ are on the probability scale $[0, 1]$ after inverse-logit transformation.
To allow heterogeneous precision across items, item-level precision parameters are modelled hierarchically:
$$z_{\phi,i} \sim \text{Normal}(0, 1)$$
$$\phi_i = \exp(\log \phi_{\text{group}} + 0.5 \cdot z_{\phi,i})$$
where the factor 0.5 is a fixed log-scale standard deviation governing item-level precision heterogeneity. The key computational optimisation lies in the likelihood specification. By the Central Limit Theorem, sample means of bounded continuous observations are approximately normally distributed. The model therefore places a Normal likelihood on observed sample means $\bar{y}_i$ rather than on individual observations:
$$\bar{y}_i \sim \text{Normal}\left(\mu_i, \sqrt{\frac{\tilde{\mu}_i(1-\tilde{\mu}_i)}{\phi_i \cdot n_i}}\right)$$
where $\tilde{\mu}_i = \text{clip}(\mu_i, 0.01, 0.99)$ prevents degenerate variance at boundary values, and $n_i$ is the number of observations contributing to item $i$'s mean. This aggregated likelihood provides an approximately $20\times$ speedup compared to modelling individual observations. For example, in a grouping with 50 items each contributing 20 observations, 50 item-level likelihood evaluations replace 1{,}000 observation-level evaluations.

The population-level mean is $M = \frac{1}{N}\sum_{i=1}^{N} \mu_i$, computed as the mean of item-level means across posterior draws (the same estimand construction as for binary inference; see Appendix~\ref{app:binary_pathway}). Posterior inference proceeds via MCMC sampling, with the credible interval for $M$ providing the basis for stopping decisions. The normalised CI width (on the $[0, 1]$ scale) is compared directly against the precision threshold \texttt{delta\_cap}; original-scale bounds (multiplied by the range $[\text{upper bound} - \text{lower bound}]$) are reported in metadata for interpretability.

\paragraph{Stopping Criteria.} Both CI width and stabilisation criteria apply to bounded continuous scores, operating on the normalised scale for consistency. The conservatism adjustment (Appendix~\ref{app:conservatism}) is applied based on normalised performance: a grouping with normalised mean score below 1\% (the default \texttt{low\_performance\_threshold}) triggers conservative stopping behaviour. The normalised CI width is multiplied by $c$ before comparison against \texttt{delta\_cap}, and the stabilisation slope threshold is divided by $c$, both extending data collection to guard against premature conclusions about near-floor performance.

\subsection{Worked Illustration}
\label{app:illustration}
To illustrate the framework specifications in Appendix~\ref{app:precision_stopping}--\ref{app:pathways}, we present a hypothetical worked example of \texttt{optstop} operating within a representative evaluation scenario. The following traces expected behaviour based on the statistical properties described in the preceding appendix subsections; empirical validation with real data appears in Section~\ref{sec:validation}.

\subsubsection{Evaluation Setup}

Consider an evaluation assessing two language models - Model A (strong performer) and Model B (weak performer, below 1\% accuracy on Task 1) - on two distinct tasks:
\begin{itemize}
    \item Task 1 (Binary): A factual question-answering task scored as correct (1) or incorrect (0)
    \item Task 2 (Ordinal): A reasoning quality assessment scored on a 0--10 rubric\footnote{Continuous scores (Appendix~\ref{app:continuous_pathway}) are also supported but omitted here for brevity.}
\end{itemize}

The ordinal groupings assume hybrid inference mode (\texttt{ordinal\_inference='hybrid'}, the default when using the package via \texttt{inspect\_ai}; standalone mode defaults to \texttt{'modal'}), which includes both modal CI and entropy convergence pathways. This configuration yields four groupings, each analysed independently (Table~\ref{tab:illustration_groupings}):

\begin{table}[htbp]
\centering \caption{Model groupings, tasks, and assumed performance characteristics} \label{tab:illustration_groupings} \begin{tabular}{l l l l} \toprule Grouping & Model & Task & Assumed Pattern \\ \midrule G1 & Model A & Task 1 (Binary) & High accuracy, consistent \\ G2 & Model A & Task 2 (Ordinal) & Scores clustered in upper range \\ G3 & Model B & Task 1 (Binary) & Low accuracy, conservatism active \\ G4 & Model B & Task 2 (Ordinal) & Low, diffuse score distribution \\ \bottomrule \end{tabular} \end{table}

Each task comprises 200 items, with up to 10 epochs permitted per item. A complete evaluation without early stopping would require 8{,}000 trials (2 models $\times$ 2 tasks $\times$ 200 items $\times$ 10 epochs).

\subsubsection{Model A: Strong Performer}

\paragraph{Grouping G1 (Binary Task):} Model A's high accuracy yields consistent success patterns across items. At the item level, items with uniform outcomes (consistent successes) achieve narrow credible intervals quickly, as the item-level Beta posterior (Appendix~\ref{app:binary_pathway}) concentrates near its upper bound. Occasional items with mixed outcomes take somewhat longer, since the posterior spans a wider range of plausible success rates.

At the grouping level, the hierarchical model rapidly accumulates evidence for a high mean success probability across items. As items contribute data, the posterior concentrates and the credible interval narrows; grouping-level precision is reassessed at discrete intervals (every \textit{X} many completed trials, where \textit{X} is set by the user via \texttt{reanalysis\_interval}). Once the grouping-level precision threshold is met, evaluation of G1 terminates - remaining items need not be evaluated, and incomplete items can be discontinued.

\paragraph{Grouping G2 (Ordinal Task):} Model A's scores cluster in the upper range of the rubric (e.g., predominantly 7--9), representing a peaked distribution. At the item level, items with consistent scoring achieve narrow modal credible intervals relatively quickly, as the Bayesian bootstrap concentrates on a clear modal category.

At the grouping level, Pathway~1 (modal CI with entropy validation) governs stopping. This requires two conditions to be satisfied independently: the hierarchical modal CI must be narrow (below \texttt{delta\_cap}), and a separate entropy check must confirm that posterior median entropy falls below the peakedness threshold (Appendix~\ref{app:ordinal_pathway}; default 80\% of $\log_2 K$ bits) - confirming genuine distributional concentration rather than a spurious peak from limited data. Grouping G2 typically achieves substantial efficiency gains: when scores cluster in a narrow range of the rubric, both conditions can be satisfied simultaneously with relatively few observations.

\subsubsection{Model B: Weak Performer}

\paragraph{Grouping G3 (Binary Task):} Model B's low accuracy triggers the conservatism regime. At the item level, two mechanisms collectively delay stopping decisions: effective CI widths are inflated by the conservatism factor $c$ (default 5) before comparison against thresholds, and adaptive prior decay slows five-fold (the prior retains influence over more epochs, maintaining wider intervals). This reflects the inferential asymmetry central to capability assessment: a string of failures does not preclude the possibility of rare successes, and such successes - if they exist - may be precisely what the evaluator seeks to detect.

Item-level stopping is therefore correspondingly rare; most items proceed through all available epochs. At the grouping level, the CI width threshold may not be achieved even with extensive data, as the conservatism mechanisms inflate the effective CI width beyond the raw posterior uncertainty. The stabilisation criterion (Appendix~\ref{app:stabilisation}) is also subject to conservatism: the slope threshold is divided by $c$ (requiring $|\hat{\beta}| < \epsilon / c$, per Appendix~\ref{app:conservatism}), making stabilisation deliberately hard to trigger. Given sufficient items, Grouping G3 eventually stops via this tightened stabilisation criterion, having established that Model B's accuracy is low - whilst affording sufficient opportunity for rare success detection.

\paragraph{Grouping G4 (Ordinal Task):} Model B's scores are both low and variable, spread across the lower portion of the rubric without a clear modal concentration. This diffuse distribution presents a different stopping dynamic. At the item level, the Bayesian bootstrap yields wide modal credible intervals, as no single category dominates. Most items proceed through all available epochs.

At the grouping level, Pathway~1 is blocked by the entropy validation gate: even if the modal CI appears narrow at some point, the high entropy signals that the distribution may not be genuinely peaked - either because it is truly spread across categories, or because limited data have not yet resolved the distributional shape. This gate prevents premature stopping based on spurious concentration.

Pathway~2 (entropy convergence) would govern stopping instead. The system evaluates whether the entropy credible interval width on the normalised $[0, 1]$ scale has fallen below a convergence threshold (default 0.10). Once this width is sufficiently narrow - indicating that posterior uncertainty about the distributional shape has been adequately resolved, even though the distribution itself is diffuse - Grouping G4 stops.
The evaluation has established that Model B produces variable, low-quality responses on this task, with the shape of this distribution (not merely a point estimate) adequately characterised.

\subsubsection{Summary of Expected Behaviour}
Several patterns emerge from this illustration:
\begin{itemize}
    \item \textbf{Performance-dependent savings}: Strong, consistent performance enables rapid posterior concentration and early stopping; weak or variable performance requires more extensive characterisation, enforced by conservatism.
    \item \textbf{Score type interactions}: Ordinal tasks may converge faster than binary tasks when performance produces a peaked distribution, as Pathway~1 can trigger once the modal CI is narrow and entropy is confirmed low. For weak performers, both score types require extensive data, though for different reasons - binary tasks due to the difficulty of precisely estimating small probabilities, ordinal tasks due to distributional diffuseness.
    \item \textbf{Pathway differentiation}: Peaked distributions stop via Pathway~1; diffuse distributions stop via Pathway~2. This routing occurs automatically based on observed data.
    \item \textbf{Adaptive scheduling}: In a live \texttt{inspect\_ai} evaluation, groupings that reach stopping criteria are marked complete and their remaining trials skipped, naturally concentrating computation on groupings that remain active (Appendix~\ref{app:groupings}) - maximising expected information gain across outstanding trials.
\end{itemize}

\section{Package Validation}

The \texttt{optstop} package (v0.4.0) was validated through a suite of integration tests conducted within the \texttt{inspect\_ai} framework (v0.3.170). This appendix summarises the validation evidence organised by the property being tested. All tests used live LLM inference (Claude Sonnet 4.5, GPT-4o, GPT-3.5 Turbo) rather than mocked outputs, ensuring that validation reflects realistic evaluation conditions including LLM response variability.

\subsection{Inference Pathway Coverage}

Twenty-nine validation runs were executed across four evaluation benchmarks spanning all three inference pathways, with all runs meeting their specified stopping criteria and producing scores consistent with benchmark expectations. These validation runs are distinct from the development test suite, which includes additional unit and stress tests with known stochastic boundary cases (e.g., MCMC timeout sensitivity, efficiency thresholds near stochastic boundaries). Table~\ref{tab:pathway_coverage} summarises the test matrix.

\begin{table}[htbp]
\centering
\caption{Inference pathway validation coverage}
\label{tab:pathway_coverage}
\begin{tabular}{l l l l}
\toprule
Pathway & Benchmarks & Scale & Modes Tested \\
\midrule
Binary & TruthfulQA, RACE-H & \{0, 1\} & Live, shadow, multi-model \\
Ordinal & WritingBench, SciKnowEval & 11-cat, 6-cat & Live, shadow \\
Continuous & WritingBench & [0, 10] & Live, shadow, multi-model \\
\bottomrule
\end{tabular}
\end{table}

Configurations ranged from 10 to 1{,}500 planned trials, with 2 to 5 epochs per item, across single-model and multi-model runs. Automatic pathway routing identified the correct inference pathway for each benchmark without manual override.

\subsection{Precision Threshold Sensitivity}

The precision threshold $\delta$ (Appendix~\ref{app:ci_widths}) governs the trade-off between estimation precision and trial savings. To validate that this parameter behaves as specified, identical evaluation configurations (TruthfulQA, RACE-H, and WritingBench; 100 items $\times$ 5 epochs each; 1{,}500 total planned trials across three independently evaluated tasks) were run under two threshold settings (Table~\ref{tab:threshold_sensitivity}).

\begin{table}[htbp]
\centering
\caption{Precision-efficiency trade-off across threshold settings}
\label{tab:threshold_sensitivity}
\small
\begin{tabular}{l c c c c c c}
\toprule
& \multicolumn{3}{c}{$\delta = 0.2$} & \multicolumn{3}{c}{$\delta = 0.05$} \\
\cmidrule(lr){2-4} \cmidrule(lr){5-7}
Task (Pathway) & Trials & Efficiency & Final CI & Trials & Efficiency & Final CI \\
\midrule
TruthfulQA (Binary) & 63 & 87.4\% & 0.163 & 238 & 52.4\% & 0.047 \\
RACE-H (Binary) & 24 & 95.2\% & 0.192 & 270 & 46.0\% & 0.049 \\
WritingBench (Continuous) & 30 & 94.0\% & $\dagger$ & 100 & 80.0\% & $\dagger$ \\
\midrule
\textbf{Overall} & \textbf{117} & \textbf{92.2\%} & & \textbf{608} & \textbf{59.5\%} & \\
\bottomrule
\multicolumn{7}{l}{\footnotesize $\dagger$ Continuous pathway CI widths reported on a normalised [0,1] scale; not directly comparable to binary posterior widths.} \\
\end{tabular}
\end{table}

In this configuration, the 4$\times$ tighter threshold required approximately 5$\times$ more trials overall (the specific multiplier depends on the underlying performance distributions). Final credible interval widths met their respective thresholds in all cases; because stopping occurs at discrete reanalysis checkpoints rather than continuously, some groupings achieved precision substantially tighter than required (e.g., TruthfulQA at 0.163 vs.\ the 0.2 threshold). At the tighter threshold ($\delta = 0.05$), higher-accuracy groupings converged faster (TruthfulQA at 238 trials vs.\ RACE-H at 270), as expected from the lower per-observation variance at extreme performance levels (Appendix~\ref{app:ci_widths}). At $\delta = 0.2$, the pattern reverses (TruthfulQA at 63 vs.\ RACE-H at 24), likely because RACE-H's CI crossed the looser threshold at an earlier reanalysis checkpoint.

Additional threshold variation tests ($\delta \in \{0.1, 0.3\}$) with binary scoring confirmed monotonic behaviour: looser thresholds produce earlier stopping with wider final intervals, while tighter thresholds produce later stopping with narrower intervals.

\subsection{Ordinal Stopping Pathway Validation}
\label{app:ordinal_pathway_validation}

The ordinal inference pathway (Appendix~\ref{app:ordinal_pathway}) was validated across two benchmarks with different scale sizes: WritingBench (11-category rubric, scores 0--10) and SciKnowEval (scores 1--5, 6 modelled categories including zero). This pairing exercises Pathway~1 (modal CI with entropy validation) under distinct distributional conditions and scale sizes.

SciKnowEval scores were near-ceiling (mean $\approx$ 4.86/5.0), producing a strongly peaked distribution with low entropy (0.43 bits vs.\ a default threshold of $0.8 \times \log_2 6 = 2.07$ bits). Pathway~1 triggered on the first grouping-level reanalysis check. As expected, the stopping point (trial 247) was identical across runs with 250 and 500 planned trials, since stopping decisions depend only on data accumulated to that point. In live (i.e., testing via \texttt{inspect\_ai}) early stopping mode, actual termination occurred at trial 259 - close to the shadow mode prediction, with the small difference attributable to in-flight trials at the time of the stopping decision.

WritingBench scores clustered in the upper range (mean $\approx$ 7.6/10) with moderate spread, producing a distribution of intermediate entropy. The scale-aware entropy threshold (Appendix~\ref{app:ordinal_pathway}) - which scales the peakedness gate proportionally to $\log_2 K$ - was necessary for Pathway~1 to trigger at this scale size; an earlier fixed-threshold implementation blocked convergence for scales above 7 categories. In shadow mode runs of 500 trials, stopping was indicated at trial 365, yielding 27\% potential efficiency.

Pathway~1 is thus validated for peaked distributions across both scale sizes, along with the scale-aware entropy threshold mechanism that enables convergence for larger ordinal scales. Pathway~2 (entropy convergence) was not triggered in these initial tests, as both benchmarks produced sufficiently peaked distributions for Pathway~1 to govern stopping. However, Pathway~2 was subsequently triggered in the matrix validation experiment (Appendix~\ref{app:downstream_equivalence}): the mid-ordinal cell (WritingBench, $K = 11$, \texttt{max\_tokens}~=~500) produced a genuinely spread distribution across the 11 categories, with entropy at 79.6\% of maximum and a mean score of $\approx 4.2/10$ - consistent with a mid-performance evaluation where responses receive a range of quality ratings rather than concentrating at any single score. No single category dominated. Entropy (79.6\% of maximum) fell just below Pathway~1's 80\% peakedness gate - so the gate did not block Pathway~1 - but the absence of a clear mode meant the modal CI width remained at 0.100, well above $\delta = 0.05$, preventing Pathway~1 from triggering. This borderline case illustrates the complementary role of Pathway~2: the entropy gate alone does not distinguish between distributions with clear modes and those that are spread but marginally below the gate threshold. Pathway~2 correctly determined that this distributional diffuseness was characterised with sufficient stability to stop (entropy CI width below the 0.10 convergence threshold), triggering at trial 857 with 57.2\% efficiency and a score deviation of just 0.001 from the full-run estimate.

\subsection{Shadow Mode Counterfactual Validation}

The shadow mode - which executes all planned trials while internally tracking when stopping criteria are met - provides a mechanism for validating stopping decisions against ground truth from complete runs. Shadow mode was used to verify two properties.

First, \emph{cross-run consistency}: a continuous-pathway configuration that achieved 87.6\% efficiency in live early stopping mode (terminating at 31 of 250 planned trials) was run separately in shadow mode. The shadow run identified a potential stopping point at trial 20, with 92\% potential efficiency. Because LLM non-determinism produces different scores across runs, the stopping points are not directly comparable; the difference reflects input variability rather than algorithmic inconsistency. The score estimates (7.536 shadow, 7.568 live) fell within the range of WritingBench means across all test configurations (7.47--7.65). A rigorous paired assessment - comparing early-stopped estimates against complete-run estimates from identical data - is presented in Appendix~\ref{app:downstream_equivalence}.

Second, \emph{diagnostic completeness}: shadow mode recorded the trial count at which each grouping's stopping criteria were first satisfied, the stopping pathway invoked (1: CI width precision; 2: CI width/entropy stability), and the full stabilisation history. These diagnostics enable post-hoc analysis of stopping behaviour without the confound of actual early termination.

\subsection{Reproducibility}

\paragraph{MCMC determinism.} Given identical input scores, the same random seed, and identical software and hardware configurations, \texttt{optstop} produces identical posterior distributions, credible intervals, and stopping decisions. This was verified by comparing paired runs that received different LLM responses (due to generation non-determinism at temperature $> 0$) but identical MCMC seeds. The resulting posteriors diverged only where input scores differed, confirming that all internal randomness is controlled by the user-specified seed.

Full end-to-end reproducibility of stopping decisions additionally requires deterministic LLM outputs (e.g., temperature = 0), which is outside the scope of the stopping framework. This distinction is important for interpreting validation results: trial count differences between nominally identical runs reflect LLM response variability, not instability in the stopping algorithm. Score estimates and credible interval widths remain consistent across runs despite this variability - for example, across all WritingBench runs the mean score ranged from approximately 7.5 to 7.6 regardless of the specific stopping point.

\paragraph{Deployment reproducibility.} To verify packaging integrity, the package was uninstalled and reinstalled from its repository (same commit, same dependency versions) before re-running the full three-benchmark validation suite. All tasks converged at final CI widths meeting the specified threshold, with scores consistent across runs (e.g., TruthfulQA 97--98\%, RACE-H 90--93\%, WritingBench $\approx$ 7.5). Trial count variance between the fresh-install run and the prior run (521 vs.\ 608 of 1{,}500 trials) was within the range attributable to LLM non-determinism.

\subsection{Logit-Normal vs.\ Beta-Binomial Model Comparison}
\label{app:model_comparison}

To justify the logit-normal parameterisation for the binary pathway (Appendix~\ref{app:binary_pathway}), we conducted a controlled simulation comparing the logit-normal hierarchical model against the conjugate Beta-Binomial alternative. Data were generated from a Beta-Binomial process - deliberately favouring that model - to provide a conservative test of logit-normal adequacy.

\paragraph{Design.} The data-generating process drew item-level success probabilities $\theta_i \sim \text{Beta}(\mu\kappa, (1-\mu)\kappa)$ with concentration $\kappa = 10$ (moderate heterogeneity), then sampled $s_i \sim \text{Binomial}(10, \theta_i)$ for each item. The parameter grid crossed 9 true population probabilities $\mu \in \{0.0, 0.05, 0.1, 0.3, 0.5, 0.7, 0.9, 0.95, 1.0\}$ with 3 sample sizes $n \in \{20, 50, 100\}$ items, yielding 27 conditions. Each condition was replicated 25 times (675 total runs). Both models used 4 chains $\times$ 1{,}000 post-warmup draws with \texttt{target\_accept}$= 0.95$, matching the \texttt{optstop} package defaults. Point estimates were posterior means of the population probability; credible intervals were 94\% HDI, matching the matrix experiment configuration (the package default is 97\%).

\paragraph{Boundary conditions.} At exact boundaries ($\mu = 0.0$ or $1.0$), the Beta distribution is degenerate and all items share identical success probability (zero between-item heterogeneity). These conditions test model behaviour under a qualitatively different regime from mid-range cases where items genuinely vary around the population mean.

\paragraph{Results.} Table~\ref{tab:model_comparison} summarises results aggregated into three regions. In the mid-range ($\mu = 0.1$--$0.9$, 15 conditions), the two models are statistically indistinguishable: absolute bias, CI width, and coverage are identical to reported precision. Near boundaries ($\mu = 0.05, 0.95$), performance remains closely matched. Both models exhibit coverage below the nominal 94\% level in non-boundary conditions (approximately 80\%), reflecting finite-sample calibration under moderate heterogeneity and the indirect nature of the population-mean estimand; crucially, coverage is closely matched between models, confirming that the logit-normal does not sacrifice calibration. At exact boundaries, both models exhibit zero coverage - an expected consequence of the degenerate data-generating process rather than a model deficiency.

The critical differentiator is computational reliability. The Beta-Binomial produced 120$\times$ more divergences than the logit-normal in the mid-range, rising to 192$\times$ near boundaries and 9{,}206$\times$ at exact boundaries (73{,}646 vs.\ 8 total divergences). Beta-Binomial effective sample sizes dropped as low as 7 at boundaries (vs $>$4{,}600 for the logit-normal at boundaries; $>$3{,}700 across all conditions), indicating severely compromised posterior exploration. These sampling difficulties stem from the Beta-Binomial's concentration parameter creating a poorly conditioned posterior geometry when data are sparse or boundary-adjacent - precisely the conditions an adaptive stopping framework must handle robustly.

\begin{table}[htbp]
\centering
\caption{Logit-normal (L-N) vs.\ Beta-Binomial (B-B) hierarchical model comparison under a Beta-Binomial data-generating process. Metrics averaged across sample sizes (20, 50, 100 items) and 25 replications per condition. Divergence ratio reports total B-B divergences divided by total L-N divergences within each region.}
\label{tab:model_comparison}
\small
\begin{tabular}{@{}l cc cc cc c@{}}
\toprule
 & \multicolumn{2}{c}{Abs.\ Bias} & \multicolumn{2}{c}{CI Width} & \multicolumn{2}{c}{Coverage} & Diverg. \\
\cmidrule(lr){2-3} \cmidrule(lr){4-5} \cmidrule(lr){6-7} \cmidrule(lr){8-8}
Region & L-N & B-B & L-N & B-B & L-N & B-B & Ratio \\
\midrule
Mid-range (0.1--0.9) & 0.022 & 0.022 & 0.070 & 0.070 & 0.80 & 0.81 & 120$\times$ \\
Near-boundary (0.05, 0.95) & 0.012 & 0.012 & 0.039 & 0.038 & 0.78 & 0.78 & 192$\times$ \\
Boundary (0.0, 1.0) & 0.007 & 0.005 & 0.011 & 0.011 & 0.00 & 0.00 & 9{,}206$\times$ \\
\bottomrule
\end{tabular}
\end{table}

\paragraph{Implications for stopping behaviour.} Because \texttt{optstop} triggers grouping-level stopping when CI width falls below \texttt{delta\_cap}, equivalent CI widths imply equivalent stopping times - confirming that the model choice does not alter the framework's adaptive behaviour in the mid-range where most evaluations operate. The logit-normal's dramatically superior sampling reliability (approximately 3$\times$ higher ESS in the mid-range, rising to over 37$\times$ at boundaries; orders of magnitude fewer divergences) ensures that posterior estimates remain trustworthy across the full performance range, including the boundary-adjacent conditions that the conservatism mechanism (Appendix~\ref{app:conservatism}) is designed to protect. Simulation code is available at \url{https://github.com/UKGovernmentBEIS/optstop}.

\subsection{Validation Scope}

The tests in this appendix and the matrix experiment (Appendix~\ref{app:downstream_equivalence}) exercise all three inference pathways and both ordinal stopping pathways. Four additional validation exercises are included: (i) a controlled conservatism sensitivity analysis across calibrated performance levels (Appendix~\ref{app:conservatism_validation}); (ii) a dedicated CI stabilisation test confirming the stabilisation criterion triggers as the primary stopping mechanism under relaxed parameters (Appendix~\ref{app:stabilisation_validation}); (iii) a presentation-order robustness analysis (Appendix~\ref{app:order_robustness}); and (iv) a fixed-$n$ baseline establishing the convergence trajectory of non-adaptive uniform sampling (Appendix~\ref{app:fixed_n_baseline}). These analyses are conducted with synthetic data (conservatism, stabilisation) or replayed shadow datasets (consistency, fixed-n), complementing the live-evaluation validation in the matrix experiment.

\subsection{Downstream Analysis Equivalence}
\label{app:downstream_equivalence}

This section provides the full experimental protocol and supplementary analyses for the $3 \times 3$ matrix experiment summarised in Section~\ref{sec:validation}. Each cell was executed in shadow mode with within-shadow comparison to isolate the pure truncation effect (see Section~\ref{sec:validation} for the rationale and design overview).

\subsubsection{Experimental Design}
\label{app:matrix_design}

Table~\ref{tab:matrix_config} details the configuration of each cell. The binary column employed distinct benchmarks and models to achieve natural performance variation, while the ordinal and continuous columns used the same benchmark (WritingBench~\cite{wu2025writingbench}) with Claude Sonnet 4.5, varying \texttt{max\_tokens} to modulate performance. This design enables both within-pathway ranking comparisons (ordinal and continuous columns) and cross-pathway generalisability assessment.

\begin{table}[htbp]
\centering
\caption{Cell configurations for the $3 \times 3$ matrix validation experiment. All cells used 200 items, 10 epochs, seed 42, $\delta = 0.05$, and 97\% credible intervals. Planned trials = items $\times$ epochs = 2{,}000 per cell (reduced by 20 for mid-binary due to 2 items failing to load).}
\label{tab:matrix_config}
\small
\begin{tabular}{@{}l l l l c@{}}
\toprule
Cell & Benchmark & Model & Pathway & \texttt{max\_tokens} \\
\midrule
low\_binary   & MATH Level 5          & GPT-3.5 Turbo       & Binary     & -- \\
mid\_binary   & GPQA Diamond          & GPT-4o              & Binary     & -- \\
high\_binary  & MMLU (0-shot)         & GPT-4o              & Binary     & -- \\
\addlinespace
low\_ordinal  & WritingBench          & Claude Sonnet 4.5 & Ordinal    & 50 \\
mid\_ordinal  & WritingBench          & Claude Sonnet 4.5 & Ordinal    & 500 \\
high\_ordinal & WritingBench          & Claude Sonnet 4.5 & Ordinal    & 5{,}000 \\
\addlinespace
low\_cont.    & WritingBench          & Claude Sonnet 4.5 & Continuous & 50 \\
mid\_cont.    & WritingBench          & Claude Sonnet 4.5 & Continuous & 500 \\
high\_cont.   & WritingBench          & Claude Sonnet 4.5 & Continuous & 5{,}000 \\
\bottomrule
\end{tabular}
\end{table}

The ordinal pathway uses an 11-category rubric ($K = 11$, scores 0--10) with an ordered logistic model and dual-pathway stopping: Pathway~1 requires both a narrow modal credible interval and low normalised entropy ($< 0.8 \times \log_2 K$); Pathway~2 requires the scaled entropy CI width to fall below 0.10 on a $[0,1]$ scale. The continuous pathway uses a hierarchical logit-normal model (Appendix~\ref{app:continuous_pathway}) with the same precision threshold ($\delta = 0.05$). The binary pathway uses a hierarchical logit-normal model (Appendix~\ref{app:binary_pathway}).

\subsubsection{HiBayES Validation Framework}
\label{app:hibayes_framework}

To assess whether truncation preserves downstream statistical conclusions, we applied five complementary analyses using hierarchical Bayesian estimation~\cite{luettgau2025hibayes}. All MCMC fits used 4 chains with 2{,}000 posterior draws per chain (1{,}000 warmup iterations each; 8{,}000 total draws), seed 42, and 94\% highest density intervals (HDIs). In each analysis, the two conditions compared are the \emph{full run} (all 2{,}000 trials) and the \emph{truncated run} (only the trials up to the would-have-stopped point), both drawn from the same shadow evaluation.

\begin{description}
\item[Analysis A] \emph{Item-matched paired comparison (primary).} For each cell, per-item epoch-averaged scores from the full run and the truncated run are paired by item identity, and a Normal model is fitted to the item-level differences. This is the primary equivalence test, as it controls for item-level variance and directly estimates the mean bias introduced by truncation.

\item[Hierarchical meta-analysis] \emph{Multiplicity-adjusted pooling.} A Bayesian hierarchical random-effects model pools the cell-level results from Analysis~A, yielding a single joint estimate of the overall truncation effect and resolving individual undecided verdicts via principled shrinkage.

\item[Analysis B] \emph{Ranking preservation (ordinal, supplementary).} Hierarchical ordered logistic regression across the three ordinal cells (\texttt{max\_tokens} $\in \{50, 500, 5{,}000\}$), fitted separately to full-run and truncated data. Verifies that relative performance rankings are preserved.

\item[Analysis C] \emph{Ranking preservation (continuous, supplementary).} Hierarchical continuous model across the three continuous cells, fitted separately to full-run and truncated data. Verifies ranking preservation with cleanly converging models.

\item[Analysis D] \emph{Per-cell posterior comparison (auxiliary).} Independent model fits to full-run and truncated data for each cell. Compares posterior distributions via ROPE testing. This analysis does not control for item-level pairing and thus has lower statistical power than Analysis~A; it serves as a consistency check.
\end{description}

\subsubsection{Item-Matched Paired Comparison (Analysis A)}
\label{app:analysis_a}

Analysis~A is the primary equivalence test. For each cell, per-item epoch-averaged scores from the full run and the truncated run are paired by item identity. A Normal model is fitted to the vector of item-level differences, yielding a posterior distribution for the mean difference $\mu_{\text{diff}}$. The 94\% HDI of $\mu_{\text{diff}}$ is compared against a ROPE of $\pm 0.02$ (binary and continuous) or $\pm 0.10$ (ordinal). Because both conditions are drawn from the same evaluation run, item-level differences reflect only the truncation effect, free from between-run variance.

Table~\ref{tab:analysis_a_detail} reports the full results. All nine models converged cleanly ($\hat{R} = 1.0$, 0 divergences, bulk ESS $> 4{,}600$ for $\mu_{\text{diff}}$ and $> 2{,}500$ for $\sigma_{\text{diff}}$ across all cells). Of the nine cells, six accept the null hypothesis of equivalence; low-binary, mid-binary, and high-ordinal return undecided verdicts. No cell rejects the null.

\begin{table}[htbp]
\centering
\caption{Item-matched paired comparison results (Analysis~A, within-shadow). $n$ = number of matched items; $\bar{d}$ = observed mean difference (full $-$ truncated); $\mu_{\text{diff}}$ = posterior mean; HDI = 94\% highest density interval; ROPE = equivalence decision. All models converged with $\hat{R} = 1.0$, 0 divergences, and bulk ESS $> 4{,}600$ for $\mu_{\text{diff}}$ and $> 2{,}500$ for $\sigma_{\text{diff}}$.}
\label{tab:analysis_a_detail}
\small
\begin{tabular}{@{}l c r r l c@{}}
\toprule
Cell & $n$ & $\bar{d}$ & $\mu_{\text{diff}}$ & 94\% HDI & ROPE \\
\midrule
low\_binary    & 200 & $+$0.006 & $+$0.007 & [$-$0.010, $+$0.025] & \textit{undecided} \\
mid\_binary    & 198 & $-$0.011 & $-$0.011 & [$-$0.029, $+$0.007] & \textit{undecided} \\
high\_binary   & 200 & $+$0.005 & $+$0.005 & [$-$0.003, $+$0.012] & accept \\
\addlinespace
low\_ordinal   & 200 & $+$0.000 & $+$0.000 & [$-$0.016, $+$0.017] & accept \\
mid\_ordinal   & 200 & $+$0.005 & $+$0.006 & [$-$0.018, $+$0.029] & accept \\
high\_ordinal  & 152 & $-$0.060 & $-$0.059 & [$-$0.114, $-$0.001] & \textit{undecided} \\
\addlinespace
low\_cont.     &  54 & $+$0.001 & $+$0.001 & [$-$0.002, $+$0.005] & accept \\
mid\_cont.     & 142 & $+$0.000 & $+$0.000 & [$-$0.004, $+$0.005] & accept \\
high\_cont.    & 100 & $+$0.003 & $+$0.003 & [$-$0.002, $+$0.009] & accept \\
\bottomrule
\end{tabular}
\end{table}

The number of matched items varies across cells because the stopping point determines how many items appear in the truncated data. Continuous cells, which achieve the highest efficiency (93--97\%), have the fewest matched items (54--142 of 200), yet the narrow within-item variance in continuous scores yields tight posteriors and clear acceptance. Most ordinal and binary cells match the full 200 items (or 198 for mid-binary, where 2 items failed to load during evaluation); however, the high-ordinal cell's early stopping at trial 152 (92.4\% efficiency) means only 152 items appear in the truncated data.

The three undecided cells merit discussion. The two binary cells - low-binary (performance: $\hat{p} \approx 0.05$, stopping at trial 318) and mid-binary (performance: $\hat{p} \approx 0.50$, stopping at trial 815) - have modest posterior mean differences ($\mu_{\text{diff}} \approx {+}0.007$ and ${-}0.011$ respectively), but per-item standard deviations of 0.135 and 0.136. The undecided verdicts do not reflect a truncation problem but rather the inherent measurement challenge of binary scoring: each observation carries at most 1~bit of information, inflating the posterior width of item-level differences. At a ROPE width of $\pm 0.05$ (matching $\delta$), both cells accept equivalence (Appendix~\ref{app:rope_sensitivity}). The high-ordinal cell shows a larger effect ($\mu_{\text{diff}} \approx {-}0.059$, 94\% HDI $[{-}0.114, {-}0.001]$), driven by a known estimand mismatch: the \texttt{optstop} ordinal pathway tracks the modal category estimate for stopping decisions, while Analysis~A evaluates mean scores. The high per-item variance ($\sigma_{\text{diff}} = 0.379$) reflects the wide spread of scores within this cell. The hierarchical meta-analysis (Appendix~\ref{app:hierarchical_meta}) resolves all three undecided verdicts via principled shrinkage, pulling the high-ordinal cell's effect toward the negligible group mean.

\begin{figure}[t]
\centering
\includegraphics[width=0.75\columnwidth]{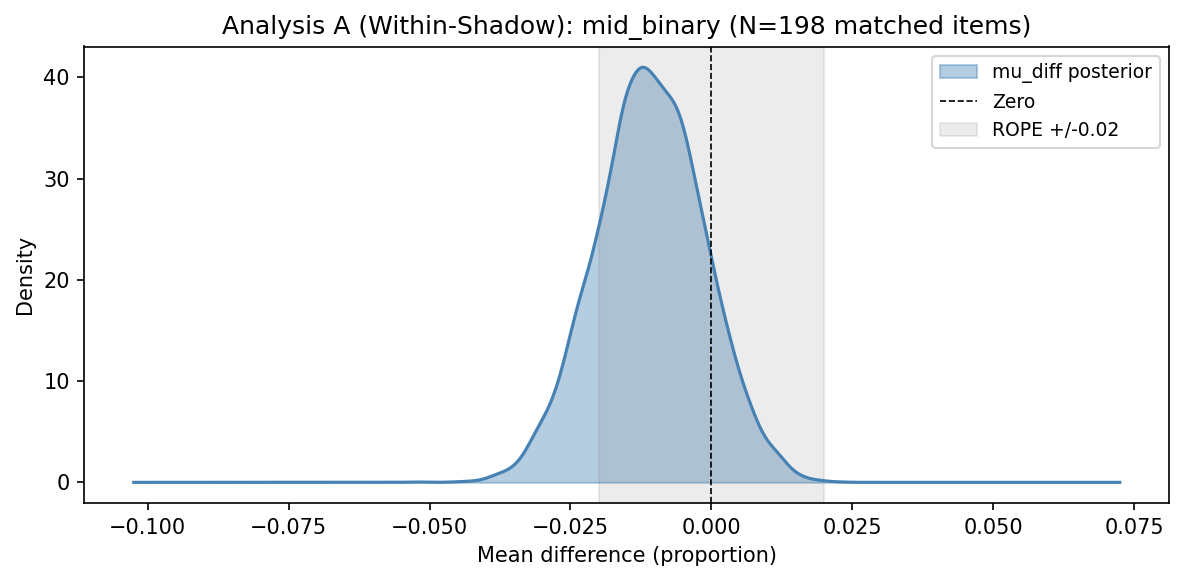}
\caption{Posterior distribution of $\mu_{\text{diff}}$ for the mid-binary cell (Analysis~A). The grey shaded band marks the ROPE ($\pm 0.02$); the dashed vertical line marks zero. The posterior is centred near $-0.011$, with the 94\% HDI $[-0.029, +0.007]$ extending beyond the ROPE due to the high per-item Bernoulli variance of this near-50\% binary benchmark, yielding an undecided verdict.}
\label{fig:analysis_a_mid_binary}
\end{figure}

Figure~\ref{fig:analysis_a_mid_binary} shows the posterior distribution for the mid-binary cell, illustrating the undecided verdict: the posterior mass is concentrated near $-0.011$ but the tails extend beyond the $\pm 0.02$ ROPE due to the high per-item variance.

\subsubsection{ROPE Sensitivity Analysis}
\label{app:rope_sensitivity}

The ROPE width is a researcher-specified parameter reflecting the magnitude of difference deemed practically negligible. To assess the sensitivity of our conclusions to this choice, we evaluated Analysis~A decisions across six ROPE widths: $\pm 0.005$, $\pm 0.01$, $\pm 0.02$, $\pm 0.05$, $\pm 0.10$, and $\pm 0.20$.

\begin{figure}[t]
\centering
\includegraphics[width=\columnwidth]{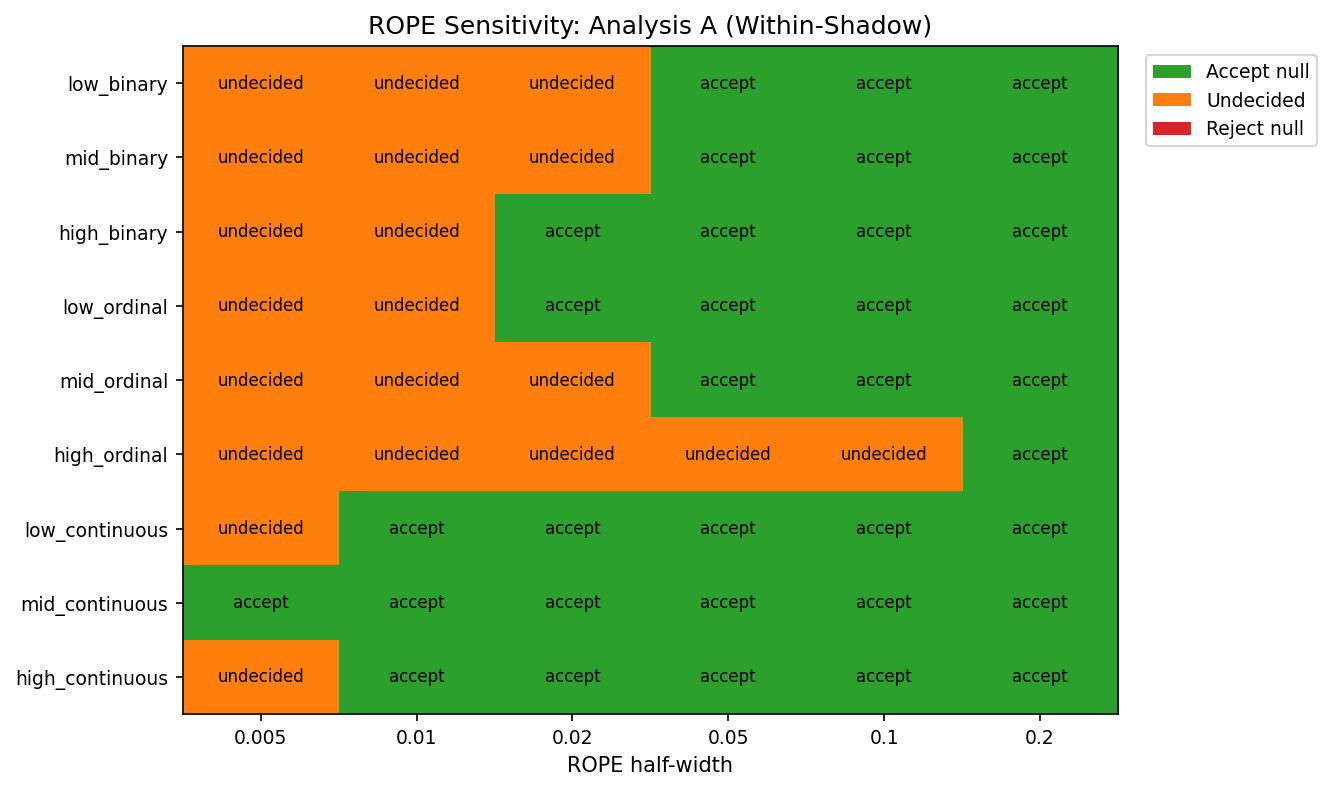}
\caption{ROPE sensitivity analysis across six ROPE widths (Analysis~A, within-shadow). Each row is a cell; columns are ROPE widths. Green: accept; orange: undecided. No cell rejects at any width. At the pathway-specific primary widths ($\pm 0.02$ for binary and continuous, $\pm 0.10$ for ordinal), 6 of 9 cells accept. At $\pm 0.05$ (matching the framework's precision threshold $\delta$), 8 of 9 cells accept; high-ordinal accepts only at $\pm 0.20$.}
\label{fig:rope_sensitivity}
\end{figure}

Figure~\ref{fig:rope_sensitivity} shows the results. At the tightest ROPE ($\pm 0.005$), most cells return undecided verdicts, as even small posterior uncertainty prevents the HDI from fitting within such a narrow region. As the ROPE widens, cells progressively accept equivalence in a staircase pattern: mid-continuous accepts at $\pm 0.005$; low-continuous and high-continuous additionally accept at $\pm 0.01$; high-binary and low-ordinal additionally accept at $\pm 0.02$; and low-binary, mid-binary, and mid-ordinal accept at $\pm 0.05$. At $\pm 0.05$ - matching the framework's precision threshold $\delta$ - eight of nine cells accept. The high-ordinal cell, with its larger effect size ($\mu_{\text{diff}} \approx {-}0.059$), accepts only at $\pm 0.20$. No cell rejects the null at any ROPE width tested.

The staircase pattern is informative: cells that accept at narrower ROPE widths are those where per-item variance is low enough to yield a tight posterior (e.g., continuous cells with real-valued scores). The two binary cells (low-binary and mid-binary) exhibit high per-item variance due to the low information content of binary observations. Low-ordinal and mid-ordinal accept at intermediate widths, while the high-ordinal cell - with its larger truncation effect driven by the modal-vs-mean estimand mismatch - accepts only at the widest ROPE tested.

\subsubsection{Hierarchical Meta-Analysis (Multiplicity-Adjusted)}
\label{app:hierarchical_meta}

The cell-level equivalence tests in Analysis~A (Appendix~\ref{app:analysis_a}) evaluate each cell independently. While the HDI+ROPE framework does not have a fixed type-I error rate in the frequentist sense, testing multiple cells independently increases the opportunity for at least one misleading verdict. To address this multiplicity concern, we fitted a Bayesian hierarchical meta-analysis that pools information across all nine cells, yielding a single joint estimate of the overall truncation effect.

\paragraph{Model specification.} We use a Normal-Normal random-effects model~\cite{gelman2013bayesian} with known cell-level standard errors:
\begin{align*}
\mu &\sim \mathrm{Normal}(0,\, 0.1) \\
\tau &\sim \mathrm{HalfNormal}(0.05) \\
\eta_i &\sim \mathrm{Normal}(\mu,\, \tau) \qquad i = 1, \ldots, 9 \\
d_i \mid \eta_i &\sim \mathrm{Normal}(\eta_i,\, \mathrm{se}_i)
\end{align*}
where $d_i$ is the observed mean difference (full run $-$ truncated, normalised to $[0,1]$) for cell~$i$, $\mathrm{se}_i = \sigma_i / \sqrt{n_i}$ is the standard error computed from the raw observed standard deviation and matched-item count from Analysis~A, $\eta_i$ is the true cell-level truncation effect, $\mu$ is the overall mean effect across all cells, and $\tau$ captures between-cell heterogeneity. Ordinal scores are divided by 10 (the maximum rubric score) to normalise to an approximately $[0,1]$ scale, consistent with the \texttt{optstop} package's internal convention. The nine cells are treated as exchangeable: while they span different inference pathways and performance levels, the model accommodates any systematic variation in truncation effects across cells through the heterogeneity parameter $\tau$, which is estimated from the data rather than assumed to be zero. A uniform ROPE of $\pm 0.02$ is applied on the normalised scale; this matches the binary and continuous ROPE from Analysis~A, and is narrower than the ordinal ROPE ($\pm 0.10$), making the joint test more stringent for ordinal cells.

The model was implemented in PyMC~\cite{abril-pla2023pymc} using non-centred parameterisation ($\eta_i = \mu + \tau \cdot z_i$, $z_i \sim \mathrm{Normal}(0,1)$) to avoid funnel-shaped posterior geometries common in hierarchical models when $\tau$ is small. Inference used NUTS sampling with 4 chains $\times$ 4{,}000 draws (2{,}000 warmup), \texttt{target\_accept}~=~0.95, and seed 42. Convergence was satisfactory: $\hat{R} = 1.00$ for all parameters, effective sample sizes exceeding 4{,}300 for $\mu$ and 3{,}500 for $\tau$, with 11 post-warmup divergences out of 16{,}000 total draws (0.07\%).

\paragraph{Results.} The posterior for the overall mean truncation effect is $\hat{\mu} = {+}0.0003$ (posterior mean), with a 97\% HDI of $[{-}0.002,\, {+}0.003]$ - entirely contained within the ROPE of $\pm 0.02$, yielding a clear \emph{accept-null} verdict. The estimated between-cell heterogeneity is $\hat{\tau} = 0.001$ (posterior median), indicating that the true cell-level effects are highly homogeneous: any truncation bias is not only negligible on average but also consistent across inference pathways and performance levels.

Figure~\ref{fig:hierarchical_meta} (Section~\ref{sec:equivalence}) presents the combined results. Panel~(a) shows the forest plot with shrinkage: grey squares indicate the original independent estimates (with Wald 95\% intervals from the observed SE), while coloured circles show the hierarchical estimates after partial pooling (with 97\% HDIs). Dashed arrows indicate the direction and magnitude of shrinkage. The most dramatic shrinkage occurs for the high-ordinal cell, whose independent estimate of $d \approx {-}0.006$ (equivalently ${-}0.060$ on the raw 0--10 ordinal scale, the source of its ``undecided'' verdict in Analysis~A) is pulled toward the group mean after pooling. The two binary cells with undecided verdicts also exhibit shrinkage, though their independent estimates ($d \approx {+}0.006$ for low-binary, $d \approx {-}0.011$ for mid-binary) are already closer to zero. This reflects the hierarchical model's partial-pooling logic: cells with larger standard errors receive greater shrinkage toward the group-level estimate. Panel~(b) shows the posterior density of $\mu$, confirming that the overall effect is negligible and falls well within the ROPE.

All nine cells receive accept-null verdicts after partial pooling, compared to 6/9 (with low-binary, mid-binary, and high-ordinal undecided) under the independent Analysis~A framework. This resolution of the three undecided cells is methodologically principled rather than a statistical artefact: the hierarchical model correctly identifies that these truncation biases are consistent with elevated per-item variance, and borrows strength from the six other cells that unanimously show negligible effects.

\paragraph{Prior sensitivity.} To verify that results are data-driven rather than prior-determined, we repeated the analysis under four prior configurations: the default ($\mu \sim \mathrm{Normal}(0, 0.1)$, $\tau \sim \mathrm{HalfNormal}(0.05)$), a wider prior ($\sigma_\mu = 0.5$), a tighter heterogeneity prior ($\sigma_\tau = 0.02$), and a wider heterogeneity prior ($\sigma_\tau = 0.10$). All four configurations produced identical ROPE decisions, and the posterior estimates of $\mu$ and $\tau$ varied by less than $10^{-4}$ across configurations. With nine cells providing information, the data dominate the priors.

\subsubsection{Ranking Preservation (Analyses B and C)}
\label{app:ranking}

As supplementary evidence, we verified that truncation preserves \emph{relative} performance comparisons between conditions - a property critical for capability ranking. We used the ordinal and continuous columns, where the same benchmark was evaluated at three \texttt{max\_tokens} levels (50, 500, 5{,}000), providing a known ground-truth ranking.

\begin{figure}[t]
\centering
\includegraphics[width=\columnwidth]{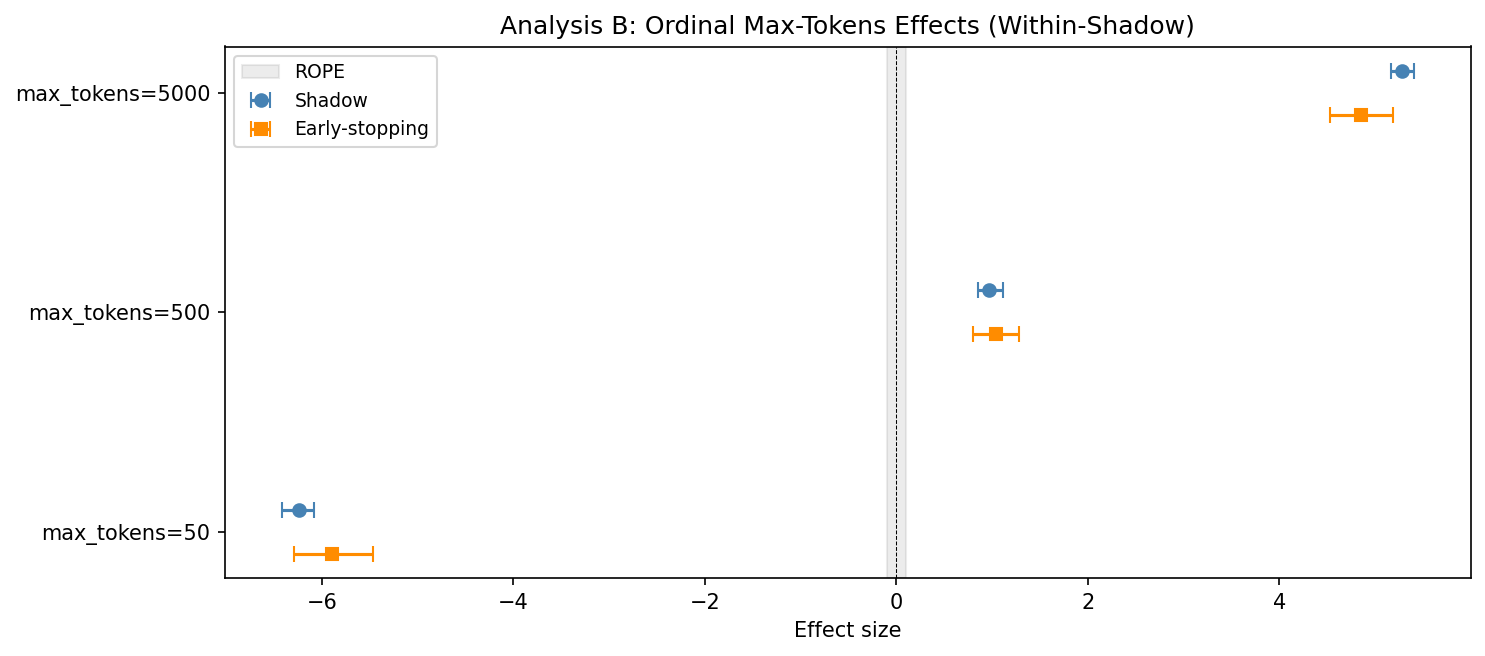}
\caption{Forest plot of model effects from hierarchical ordered logistic regression (Analysis~B). Blue: full run; orange: truncated. Error bars show 94\% HDIs. The ranking $\texttt{max\_tokens} = 5{,}000 > 500 > 50$ is preserved in both conditions, with non-overlapping HDIs between all adjacent levels.}
\label{fig:analysis_b_forest}
\end{figure}

\paragraph{Analysis B: Ordinal ranking.} Figure~\ref{fig:analysis_b_forest} shows the forest plot of model effects from the hierarchical ordered logistic regression. In both full-run and truncated conditions, the ranking $\texttt{max\_tokens} = 5{,}000 > 500 > 50$ is preserved with non-overlapping 94\% HDIs between all adjacent levels. The effect sizes are large: the posterior mean for \texttt{max\_tokens}~=~50 is $-6.24$ (full run) and $-5.89$ (truncated) on the log-odds scale, compared to $+5.28$ and $+4.85$ for \texttt{max\_tokens}~=~5{,}000. The full-run model exhibited severe convergence failure (7{,}982 of 8{,}000 post-warmup samples divergent, $\hat{R}$ up to 1.14, bulk ESS as low as 23), likely due to the high dimensionality of the 10-category ordered logistic model with 2{,}000 observations per cell. The truncated model showed improved but still imperfect convergence (332 divergences, $\hat{R} = 1.0$). The absolute posterior estimates from this analysis are therefore unreliable and should not be interpreted in isolation. We retain Analysis~B because it illustrates a practically important point: even under the challenging posterior geometry that ordinal models can produce - whether from data sparsity, high dimensionality, or distributional complexity - the ranking signal is sufficiently strong that it survives poorly-behaved MCMC. The effect sizes are separated by several standard deviations (the smallest adjacent gap exceeds 3 posterior standard deviations), and the ranking is consistent across both conditions. Analysis~C, which models the same data through the continuous pathway with clean convergence (0 divergences, ESS $> 3{,}200$), provides the confirmatory evidence for ranking preservation.

\begin{figure}[t]
\centering
\includegraphics[width=\columnwidth]{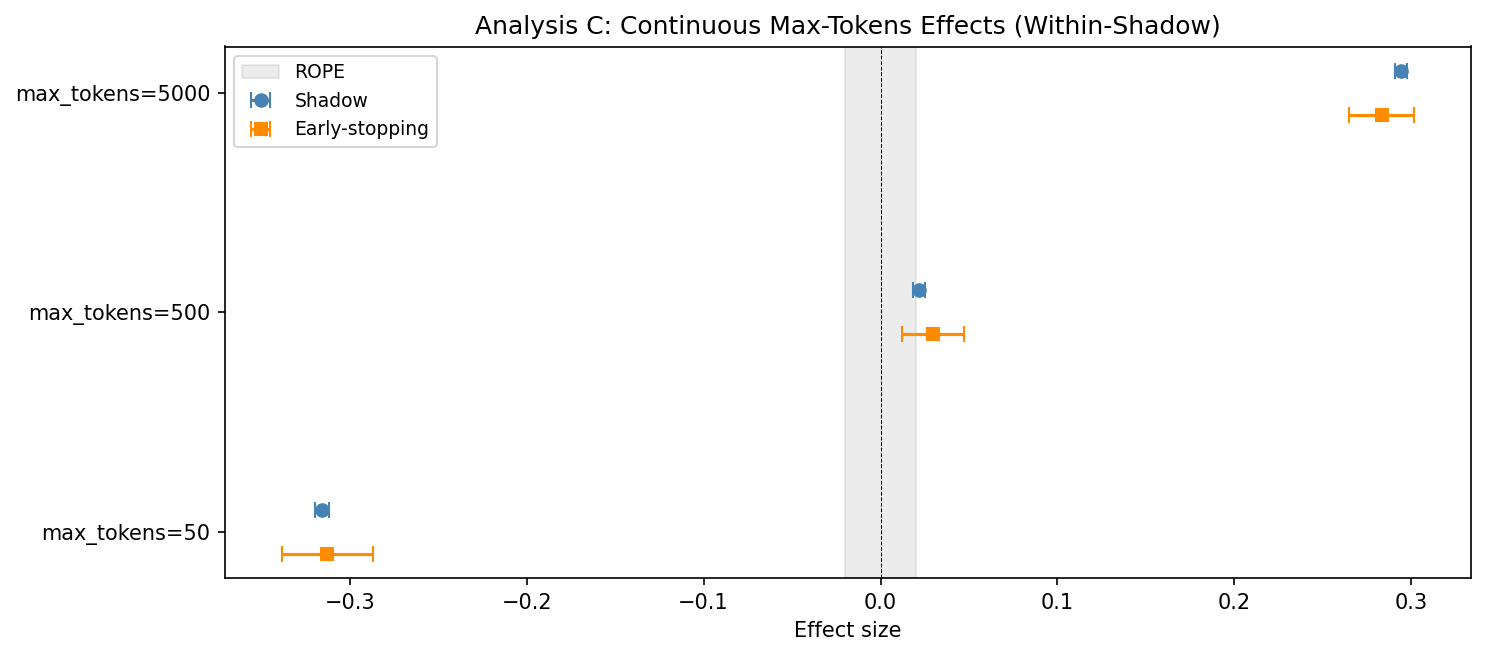}
\caption{Forest plot of model effects from hierarchical continuous model (Analysis~C). Blue: full run; orange: truncated. The ranking $\texttt{max\_tokens} = 5{,}000 > 500 > 50$ is preserved with clean convergence (0 divergences) in both conditions.}
\label{fig:analysis_c_forest}
\end{figure}

\paragraph{Analysis C: Continuous ranking.} Figure~\ref{fig:analysis_c_forest} shows the corresponding forest plot for the continuous pathway. Both full-run and truncated models converged cleanly (0 divergences, $\hat{R} = 1.0$, ESS $> 3{,}200$ for all parameters). The ranking is again preserved with non-overlapping HDIs. Posterior mean differences between full-run and truncated conditions are small: $|\Delta| \leq 0.011$ for each \texttt{max\_tokens} level. Model effects are closely aligned across conditions.

Pairwise comparisons between \texttt{max\_tokens} levels show consistent effect sizes across conditions: the gap between \texttt{max\_tokens}~=~5{,}000 and~50 is 0.610 (full run) vs.\ 0.596 (truncated); between 500 and 50, 0.338 vs.\ 0.342. All pairwise ROPE tests return ``undecided'' rather than ``reject,'' confirming that any differences between conditions are small relative to the between-level performance gaps.

\subsubsection{Per-Cell Posterior Comparison (Analysis D)}
\label{app:analysis_d}

As an auxiliary consistency check, Analysis~D fitted independent models to the full-run and truncated data for each cell and compared the resulting posterior distributions via ROPE testing. Because the posteriors are fitted independently (rather than to item-paired differences), this analysis has lower statistical power than Analysis~A. No cell rejects the null hypothesis of equivalence. The undecided verdicts in this analysis reflect the reduced sample size of the truncated condition and the absence of item-level pairing, rather than meaningful truncation bias. Analysis~A (Appendix~\ref{app:analysis_a}), which controls for item-level variance, should be preferred for equivalence assessment.

\subsubsection{Information Gain}

\begin{figure}[t]
\centering
\includegraphics[width=\columnwidth]{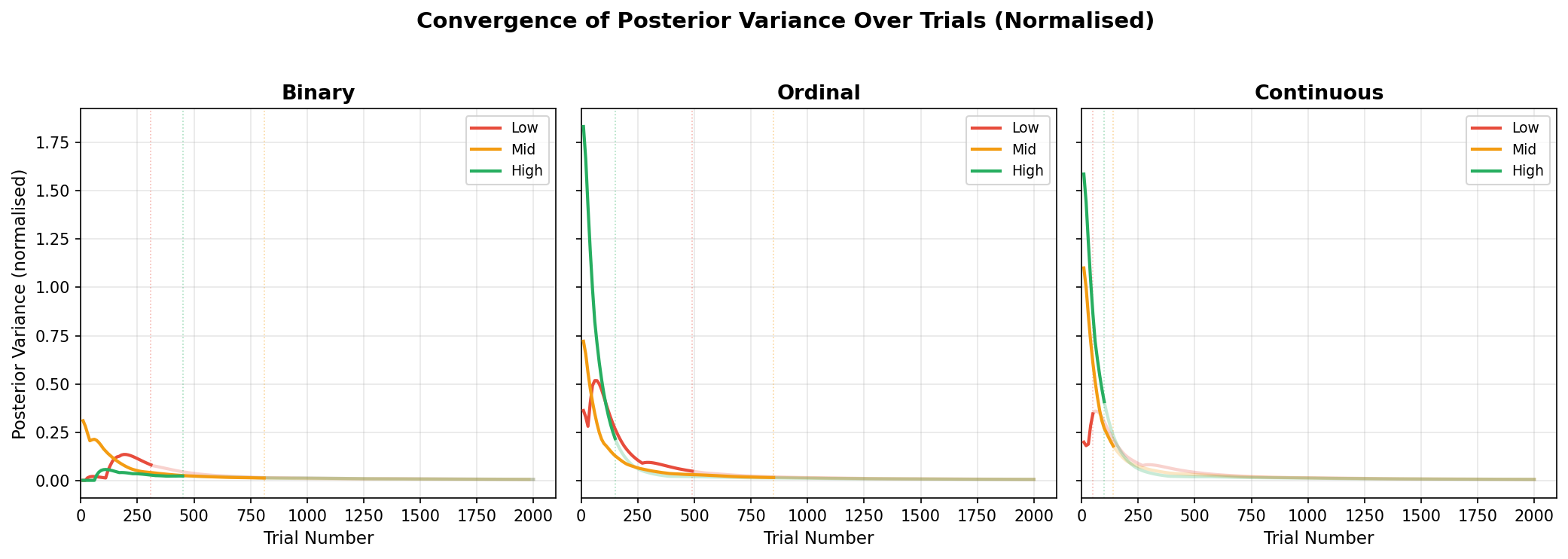}
\caption{Convergence of posterior variance over trials. Curves show the running posterior variance of each cell's group-level estimate, normalised by the cell's full-run between-item variance; vertical dotted lines indicate would-have-stopped points. Diminishing returns are evident across all pathways, with posterior variance collapsing within the first 200--400 trials.}
\label{fig:info_gain}
\end{figure}

Figure~\ref{fig:info_gain} shows the convergence of posterior uncertainty as a function of trial number for each cell: the running posterior variance of the group-level estimate, normalised by each cell's full-run between-item variance. Across all pathways, the curves exhibit diminishing returns: posterior variance collapses within the first 200--400 trials, with subsequent trials yielding progressively smaller reductions in uncertainty. The continuous cells show the steepest initial declines, consistent with their high per-observation information content and early stopping points (54--142 trials). The binary cells show the most gradual declines, consistent with the low information content of binary observations.

\subsubsection{Known Limitations}
\label{app:limitations}

Several limitations of the matrix validation design should be noted.

\begin{enumerate}
\item \emph{Epoch pseudoreplication.} Each item is evaluated across 10 epochs, and observations within an item may be correlated. The stopping framework treats epochs as exchangeable within items, which inflates the effective sample size relative to fully independent observations. This affects convergence speed but does not bias point estimates.

\item \emph{Selection bias in truncated data.} When early stopping truncates an evaluation, only a subset of items (or a subset of epochs per item) have been observed. Items evaluated early may differ systematically from later items if the item ordering is correlated with difficulty. The within-shadow comparison uses a single run with fixed item ordering, so this bias is captured directly in the truncation effect. In deployment, the stopping framework uses the same fixed ordering, so the validation is representative.

\item \emph{Multiple comparisons.} Nine cells were tested independently in Analysis~A. Although the HDI+ROPE framework does not have a fixed type-I error rate, testing multiple cells increases the opportunity for at least one misleading verdict. To address this, we fitted a hierarchical meta-analysis (Appendix~\ref{app:hierarchical_meta}) that shares information across cells via partial pooling. The joint model yields an overall accept-null verdict ($\mu = {+}0.0003$, 97\% HDI $[{-}0.002, {+}0.003]$) and resolves all three undecided results (two binary, one ordinal) via principled shrinkage.

\item \emph{Normal approximation (Analysis~A).} Analysis~A assumes that item-level differences are approximately Normal. For binary scores, the true difference distribution is discrete (epoch-averaged scores take values $k/10$ for $k \in \{0, \ldots, 10\}$, so differences are multiples of $0.1$ in $[-1, +1]$). The Normal approximation is adequate for inference about the mean ($n \geq 54$ in all cells) but may not accurately characterise the tails of the difference distribution.
\end{enumerate}

\subsection{Fixed-$n$ Baseline}
\label{app:fixed_n_baseline}

To contextualise the efficiency gains reported in Section~\ref{sec:validation}, fixed-sample-size estimates were computed at trial checkpoints (100--2{,}000 trials) across 100 random item orderings for each matrix cell. This establishes the convergence trajectory that a non-adaptive (uniform sampling) approach would follow and provides a reference for the precision achievable at each data volume.

Binary and continuous estimates converge smoothly toward their full-run values, with CI widths narrowing monotonically. At the trial counts where \texttt{optstop} triggered stopping, fixed-$n$ sample means across 100 random orderings deviate from the full-run value by at most 0.017 (binary) and 0.003 (continuous), with cross-ordering variability (standard deviation) below 0.005 and 0.008 respectively - well within the $\delta = 0.05$ precision threshold. Ordinal deviations are smaller still ($< 0.002$) but exhibit quantisation at 0.1 increments (for the 0--10 scale) because the ordinal estimand is the modal category; this is inherent to the discrete category structure rather than a computational artefact.

\subsection{Presentation-Order Robustness}
\label{app:order_robustness}

The precision-based stopping framework assumes exchangeability of observations within groupings (Appendix~\ref{app:precision_stopping}). To validate that stopping decisions are robust to presentation order, each cell's shadow dataset from the matrix experiment was replayed 15 times through \texttt{optimal\_stopping\_posthoc} with epoch-interleaved processing order (\texttt{reanalysis\_interval}$= 10$) and different item-shuffle seeds (seeds 1000--1014). Three properties were assessed using the full-run estimate as reference.

\paragraph{Results.} All 135 replications (9 cells $\times$ 15 shuffles) triggered early stopping, yielding a 100\% stop rate across all conditions. Estimate stability was high: theta standard deviation across shuffles was $< 0.006$ for binary and continuous cells, and $< 0.033$ for ordinal (the ordinal maximum is driven by mid\_ordinal, where the modal category estimate alternates between adjacent categories across shuffles). Efficiency ranges were consistent within pathway: binary 57.7--85.6\%, continuous 93.1--97.9\%, ordinal 95.7--97.3\%. The ordinal post-hoc efficiencies are substantially higher than the shadow-mode values reported in Section~\ref{sec:efficiency_fidelity} (57--92\%), because item shuffling allows the entropy criterion to be satisfied with fewer items than the fixed ordering used in the shadow run; this demonstrates that ordinal stopping efficiency is sensitive to presentation order. Binary and continuous post-hoc efficiencies closely match the shadow-mode values (within 1--6pp and $< 1$pp respectively).

Empirical CI coverage - the proportion of shuffled replications in which the early-stopped credible interval contained the full-run estimate - was 100\% across all binary and continuous cells (90/90 comparisons). This high coverage is a design property: because stopping triggers when CI width reaches $\delta = 0.05$, the resulting intervals are wide enough to absorb the observed truncation biases ($\sim$0.01). This should be interpreted as a precision-guarantee check - the stopping threshold ensures intervals are at least $\delta$-wide at termination - rather than evidence of Bayesian calibration in the frequentist sense. The probability of observing 0 misses in 90 trials under true 97\% coverage is $\sim$6.4\%, consistent with CIs that are moderately over-wide by design.

\paragraph{Ordinal coverage caveat.} Ordinal CI coverage was 0/15 (low), 12/15 (mid), and 1/15 (high). This apparent failure reflects an estimand mismatch rather than miscalibration: the ordinal pathway estimates the \emph{modal category} (most probable score), not the mean. The group-level theta for ordinal groupings represents modal\_category / max\_score, and the credible interval brackets the mode. When the score distribution is skewed, the mode and mean diverge - for example, low\_ordinal has mode $= 1/10 = 0.1$ but raw mean $= 0.11$, with the mean falling systematically outside the (correctly narrow) modal CI. Mid\_ordinal achieves partial coverage (12/15) because the modal category is uncertain at mid-performance levels, producing wider CIs that sometimes contain the mean. Pairwise CI overlap (Jaccard index) provides a more appropriate consistency metric for ordinal groupings: 0.954 (low), 0.729 (mid), 0.772 (high). The lower Jaccard for mid\_ordinal reflects instability in the modal category estimate when the score distribution is spread across multiple categories (modal estimate alternates between 0.35, 0.40, and 0.45 across shuffles, with correspondingly variable stopping points: CV $= 0.34$, items used ranging from 30 to 90).

\paragraph{Bias decomposition.} Total bias (early-stopped estimate minus full-run estimate) was decomposed into subset selection bias and model bias components. Subset selection bias was negligible for continuous ($\leq 0.002$) and small for ordinal ($\leq 0.007$) pathways. Binary subset selection was somewhat larger (up to 0.012 for low\_binary), reflecting higher item-level variance in binary scoring where which items fall in early epochs matters, but remained well within $\delta$. These results suggest that items sampled before stopping do not exhibit substantial selection bias, even when as few as 20\% of items are evaluated (ordinal and continuous pathways, where group-level stopping can trigger within the first epoch). The model bias component (reflecting HDI midpoint shift, prior shrinkage, and Bayesian versus frequentist estimand differences) dominated total bias for the continuous and ordinal pathways, varied in direction with performance level, but remained within $\delta = 0.05$ in all cases.

\subsection{Conservatism Sensitivity}
\label{app:conservatism_validation}

To isolate the conservatism mechanism (Appendix~\ref{app:conservatism}), a controlled experiment varied the conservatism factor $c \in \{1, 5, 10, 50\}$ across four calibrated performance levels (1\%, 5\%, 10\%, 15\% true success rate) using synthetic binary data (200 items $\times$ 10 epochs = 2{,}000 trials per condition). All conditions used the hierarchical PyMC binary model with \texttt{low\_performance\_threshold} $= 0.1$, ensuring conservatism activation for the 1\% and 5\% conditions (where actual performance falls below the threshold) while the 10\% and 15\% conditions serve as controls where conservatism is inactive.

\paragraph{Key findings.} At very low performance (1\%), conservatism operates as intended: $c = 1$ permits premature stopping at 120 trials with an overestimate ($\hat{\theta} = 0.027$ versus true 0.0065), while $c = 5$ delays stopping to 910 trials with a more accurate estimate ($\hat{\theta} = 0.008$). Higher values ($c = 10, 50$) consume most of the data budget (1{,}590--1{,}920 trials) for marginal accuracy gains. At moderate performance ($\geq 10\%$), conservatism has negligible effect regardless of $c$ because performance exceeds the low-performance threshold.

Two stopping pathways interact with conservatism differently. The CI-width pathway (stopping when $\text{width} \times c < \delta$) is deterministic and monotonic but requires progressively more data at higher $c$. When this pathway becomes unreachable within the data budget, stopping falls through to the slope/stabilisation pathway, which is inherently more stochastic - producing non-monotonic behaviour at high $c$ values ($c \geq 10$) and MCMC-dependent trial counts (coefficient of variation up to 13\% at $c = 50$ across repeated MCMC runs on identical data). Multi-seed validation (5 data seeds $\times$ 5 MCMC repetitions) confirmed that this non-monotonicity is systematic rather than a single-seed artefact: only 2/5 data seeds showed monotonic trial counts across $c$ values at 1\% performance.

These results informed the package defaults: $c = 5$ (reduced from 10) with \texttt{low\_performance\_threshold} $= 0.01$ (raised from 0.001) provides adequate protection against premature stopping at very low performance while keeping the effective CI-width target ($\delta / c = 0.01$) achievable with moderate data budgets. In the conservatism experiment, $c = 5$ at 1\% performance delayed stopping sufficiently to produce accurate estimates ($\hat{\theta} = 0.008$ vs.\ true 0.0065) while $c = 1$ permitted premature stopping with a fourfold overestimate; higher values ($c \geq 10$) consumed most of the data budget for marginal accuracy gains. The coupling between these parameters is documented in the package: practitioners with unusually low expected performance or large data budgets may increase $c$ accordingly.

\subsection{CI Stabilisation Validation}
\label{app:stabilisation_validation}

The CI stabilisation criterion (Appendix~\ref{app:stabilisation}) serves as a fallback stopping mechanism when CI width converges too slowly to reach $\delta$. To validate this code path in isolation, synthetic binary data were generated at three performance levels (20\%, 50\%, 80\%) with three precision thresholds ($\delta \in \{0.05, 0.01, 0.005\}$), using relaxed stabilisation parameters (\texttt{CI\_delta} $= 0.001$, 100$\times$ the default; \texttt{stab\_window} $= 10$, versus default 15) to ensure the stabilisation pathway activates.

All 9 conditions stopped via stabilisation with consistent behaviour: efficiency ranged from 75--81\% regardless of performance level or $\delta$, with theta estimates close to realised sample performance (maximum absolute deviation 0.017). The insensitivity to $\delta$ confirms that stabilisation is slope-based rather than width-based - it detects when further data collection yields diminishing returns, independent of the precision target.

\paragraph{Scope limitation.} This analysis validates that the stabilisation code path executes correctly and produces sensible results. However, under the more stringent package defaults (\texttt{CI\_delta} $= 0.00001$, \texttt{stab\_window} $= 15$), no cell in the production matrix experiment (Section~\ref{sec:validation}) would have triggered stabilisation. The stabilisation criterion functions as a safety net for scenarios where CI convergence is very slow; under typical evaluation conditions with default settings, stopping is driven by the CI-width criterion. The gap between the tested parameters (100$\times$ relaxed) and production defaults means that the precise behaviour of stabilisation under default settings remains empirically uncharacterised, though the mechanism is structurally identical.

\end{document}